\documentclass{article}

 \usepackage[main, final]{neurips_2026}

\usepackage[utf8]{inputenc} 
\usepackage[T1]{fontenc}    
\usepackage{hyperref}       
\usepackage{url}            
\usepackage{booktabs}       
\usepackage{amsfonts}       
\usepackage{nicefrac}       
\usepackage{microtype}      
\usepackage{xcolor}         
\usepackage{subcaption}
\usepackage{graphicx}
\usepackage{amsmath}
\usepackage{makecell}
\usepackage[table]{xcolor}  
\usepackage{pifont}
\usepackage{enumitem}
\usepackage{tabularx}
\usepackage{array}
\usepackage{tcolorbox}

\usepackage{titlesec}
\titlespacing*{\section}{0pt}{5pt plus 1.5pt minus 1.5pt}{5pt plus 1.5pt minus 1.5pt} 
\titlespacing*{\subsection}{0pt}{5pt plus 1.5pt minus 1.5pt}{5pt plus 1.5pt minus 1.5pt} 
\titlespacing*{\paragraph}
{0pt}    
{1em}    
{1em}    
\newcommand{\MYhref}[3][blue]{\href{#2}{\color{#1}{#3}}} 
\title{ReasonEdit: Towards Interpretable Image Editing Evaluation via Reinforcement Learning}

\author{
Honghua Chen\textsuperscript{1,2*},
Zitong Xu\textsuperscript{2*},
Huiyu Duan\textsuperscript{2†},
Xinyun Zhang\textsuperscript{1},
Xiongkuo Min\textsuperscript{2†},
Guangtao Zhai\textsuperscript{2†}
\vspace{0.2em}
\\
\textsuperscript{1}University of Electronic Science and Technology of China
\vspace{0.2em}
\\
\textsuperscript{2}Shanghai Jiao Tong University
\vspace{0.2em}
\\
\textsuperscript{*}Equal Contribution
\\
\vspace{0.2em}
\textsuperscript{†}Corresponding Authors
}

\begin{document}

\maketitle

\begin{abstract}
Recent text-guided image editing (TIE) models have achieved remarkable progress, however, many edited results still suffer from artifacts, unintended modifications, and suboptimal aesthetics. Although several benchmarks and evaluation methods have been proposed, most existing approaches rely on scalar scores and lack interpretability. This limitation largely stems from the absence of high-quality interpretation datasets for TIE and effective reward models to train interpretable evaluators. To address these challenges, we introduce \textbf{ReasonEdit-22K}, the first dataset that combines 22K edited images with 113K Chain-of-Thought (CoT) samples, along with 1.3M human judgments assessing these interpretations in terms of \textbf{logicality, accuracy, and usefulness}. Building upon this dataset, we propose \textbf{RE-Reward}, a multimodal large language model (MLLM)-based reward model designed to provide human-aligned feedback for evaluating interpretable reasoning in image editing. Furthermore, we develop \textbf{ReasonEdit}, which is trained using reward signals derived from RE-Reward and the Group Relative Policy Optimization (GRPO) algorithm to learn an interpretable evaluation model. Extensive experiments demonstrate that ReasonEdit achieves superior alignment with human preferences and exhibits strong generalization across public benchmarks. In addition, it is capable of generating high-quality interpretable evaluation text, enabling more transparent and trustworthy assessment for image editing. The code is available at \MYhref[magenta]{https://github.com/IntMeGroup/ReasonEdit}{https://github.com/IntMeGroup/ReasonEdit}.
\end{abstract}

\section{Introduction}
Text-guided image editing (TIE) enables users to modify visual content through natural language instructions \citep{brooks2023instructpix2pix}. Despite the impressive performance of state-of-the-art models (e.g., Nano-Banana \cite{nanobanana}, Seedream4.0 \cite{seedream4}, Qwen-Image-Edit \cite{qwenedit}), edited images frequently suffer from unnatural limb poses, lighting inconsistencies and unintended modifications. Existing evaluation methods primarily rely on scalar metrics \citep{lmm4edit,editreward,editscore,xu2026edithf1mmillionscalerichhuman}. While providing a convenient performance summary, they lack interpretability and fail to explain why an edit succeeds or fails, limiting their usefulness for diagnosing errors or offering actionable feedback.

Recent efforts have explored reasoning-based evaluation to address this limitation \citep{han2025unireditbench,wang2026creval}. However, progress is hindered by two key challenges: the absence of large-scale, high-quality interpretation datasets for TIE, and the lack of effective reward models to train interpretable evaluators. Existing reward models \citep{clipscore,imagereward,sarto2023positive} primarily focus on text-image alignment and are ill-suited for providing fine-grained feedback on evaluation text, making it difficult to produce human-aligned, interpretable assessments.

To bridge these gaps, we introduce \textbf{ReasonEdit-22K}, the first large-scale dataset designed for interpretable TIE evaluation. It comprises 22K edited images paired with 113K MLLM-generated Chain-of-Thought (CoT) annotations, 1.3M human judgments assessing their logicality, accuracy, and usefulness, as well as 12K human-revised high-quality CoT samples. This dataset provides comprehensive supervision for generating reliable, human-aligned reasoning.

Building upon this dataset, we develop \textbf{RE-Reward}, an MLLM-based reward model that provides fine-grained feedback on evaluation reasoning along the aforementioned dimensions. Leveraging these reward signals, we propose \textbf{ReasonEdit}, an interpretable evaluation model trained via supervised fine-tuning (SFT) and Group Relative Policy Optimization (GRPO). This paradigm enables the model to generate high-quality evaluation reasoning strongly aligned with human preferences.

Extensive experiments demonstrate that ReasonEdit achieves superior human alignment and qualitative interpretability, generalizing effectively across public benchmarks to promote transparent and diagnostically useful TIE evaluation.

The main contributions of this work include:
\begin{itemize}[left=12pt, labelsep=0.6em, labelwidth=0pt]
\item We introduce \textbf{ReasonEdit-22K}, the first large-scale dataset for interpretable TIE evaluation, featuring comprehensive CoT annotations and human judgments on reasoning quality.
\item We propose \textbf{RE-Reward}, an MLLM-based reward model that provides fine-grained, human-aligned feedback on TIE evaluation reasoning.
\item We develop \textbf{ReasonEdit}, an interpretable evaluation model trained via SFT and GRPO, capable of generating transparent and diagnostically useful evaluation text.
\item We conduct extensive experiments demonstrating ReasonEdit's superior performance in human alignment, interpretability, and generalization across public benchmarks.
\end{itemize}

\section{Related work}
\subsection{Text-guided image editing}

Text-guided image editing (TIE) has witnessed remarkable progress with the advent of large-scale diffusion models such as Stable Diffusion \cite{SD} and FLUX \cite{FLUX}. Early methods such as \textit{InstructPix2Pix} \cite{brooks2023instructpix2pix} and \textit{MagicBrush} \cite{zhang2024magicbrush} established the paradigm of instruction-following image editing on real images, providing paired data that enables supervised learning for controllable edits.
Recently, unified models \cite{qwenedit, omnigen2, dreamomni2, nanobanana, seedream4, gptimage} integrate understanding and generation within a single framework, enabling more flexible interpretation of editing instructions and producing highly realistic and diverse edits. However, this increased capability also makes evaluation more challenging \citep{xu2026edithf1mmillionscalerichhuman}, as existing methods struggle to fully capture instruction adherence, content preservation, and local changes.

\subsection{Reward model}
Reward models (RMs) serve as proxies for human intent and have evolved from scalar preference estimation toward richer, more interpretable forms of supervision. Early methods such as \textit{ImageReward} \cite{xu2023imagereward} and \textit{PickScore} \cite{kirstain2023pickscore} rely on single-value outputs for global preference ranking, but provide limited interpretability for diagnosing specific failures. To address this, recent studies explore richer signals, including visual reasoning-based evaluation \cite{lin2024vqascore}, textual critique generation \cite{zheng2024critique}, and decomposable attribution methods \cite{mo2025reward}. These efforts reflect a shift toward more explainable reward modeling with fine-grained feedback.

\subsection{Visual quality assessment}
Visual quality assessment (VQA) aims to evaluate visual content in alignment with human perception \cite{finevq,lmm4edit,harmonyiqa,wang2025lmm4lmm}. In image editing, most existing approaches \cite{lmm4edit,editreward,editscore,xu2026edithf1mmillionscalerichhuman} rely on numerical scores across one or multiple evaluation dimensions, but still lack explicit interpretability. As a result, they provide limited insight into why a generated or edited result receives certain evaluations, making error diagnosis difficult. To alleviate this issue, recent works \cite{wu2025visualquality,liu2025unlocking} incorporate reinforcement learning to enable interpretable evaluators that can generate textual feedback. However, these methods are not specifically tailored to image editing scenarios, where fine-grained evaluation of instruction adherence and content preservation is required.

\section{ReasonEdit-22K}
\label{data:all}

To address the challenges in interpretable evaluation for image editing, we introduce the first large-scale benchmark \textbf{ReasonEdit-22K}, featuring 1.5K source images and \underline{22K} image-edit pairs, accompanied by 113K MLLM-generated evaluation texts, over 1.3M fine-grained human ratings across three dimensions, and 12K expert-refined high-quality evaluation texts derived from these texts.

\subsection{Task design}

\textbf{ReasonEdit-22K} is structured into two parts: a high-quality Chain-of-Thought (CoT) text dataset \textbf{ReasonEdit-CoT-12K} and a CoT evaluation reward dataset \textbf{ReasonEdit-Reward-113K}. ReasonEdit-CoT-12K provides interpretable reasoning and scalar scores for editing quality. We evaluate these edits across three primary dimensions: \textit{visual quality}, \textit{instruction alignment}, and \textit{content preservation}. While ReasonEdit-Reward-113K focuses on evaluating the quality of the MLLM-generated CoT itself, containing expert human annotations on the critics produced by 9 distinct MLLMs. The annotations assess the CoT texts based on three criteria: 
\begin{itemize}
    \item \textbf{Logicality} assesses the soundness of its reasoning.
    \item \textbf{Accuracy} verifies it is factually correct and free of hallucinations.
    \item \textbf{Usefulness} measures whether it provides valuable, evidence-based insights.
\end{itemize}
This dual-dataset design ensures that our interpretable evaluation framework not only establishes a rigorous standard for image editing models, but also trains a reward model capable of assessing the reasoning capabilities of MLLMs in visual quality evaluation.

\begin{figure}
  \centering
  \includegraphics[width=\linewidth]{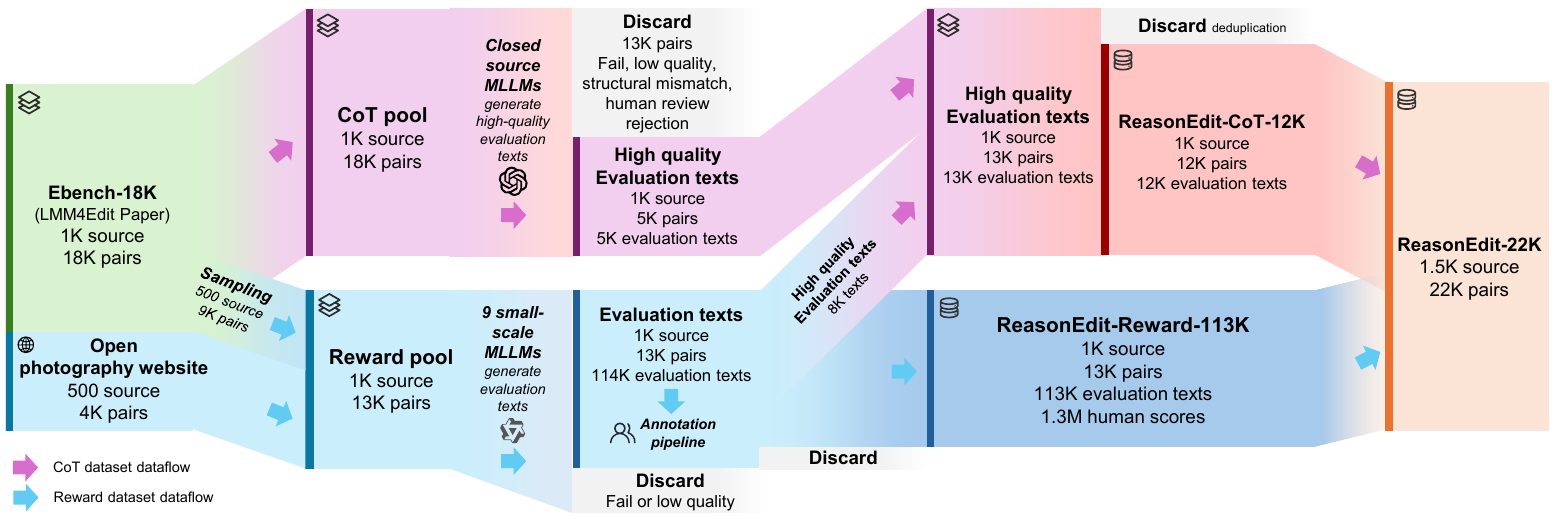}
  \caption{Dataflow during the construction of ResonEdit-22K and its two subsets}
  \label{fig:dataflow} 
\end{figure}

\subsection{Annotation pipeline}

The construction of the two subsets followed a rigorous pipeline, as demostrated in Figure \ref{fig:dataflow}. For ReasonEdit-CoT-12K, the source images and their corresponding edited counterparts are derived entirely from EBench-18K\citep{lmm4edit}. This subset encompasses 1K source images, 412 distinct editing instructions, and 18K outputs from 17 generative models, covering 8K+ pairs separated in 8 types of high-level semantic tasks, and 9K+ pairs separated in 13 types of low-level tasks.

As for the ReasonEdit-Reward-113K, we sampled 520 source images, 9K source-edited pairs from EBench-18K and additionally curated 430 high-quality source images from professional photography websites. To generate the candidate edits for the newly collected 430 images, we employed 9 state-of-the-art TIE models, such as Qwen-Image-Edit \cite{qwenedit} and NanoBanana \cite{nanobanana}, paired with 354 diverse editing instructions, covering 26 types of requirements and yield 4K source-edited pairs. Finally, we obtained $520 \times 17 + 430 \times 9 = 114,390$ source-edited pairs.

\subsubsection{CoT generation pipeline}

First, we generate candidate CoT texts with the resulting triplet above, each comprising the source image, the edited image, and the text instruction. They were processed by 9 small-scale MLLMs such as Qwen3-VL-8B \cite{qwen3}, Intern-VL-3.5-8B \cite{internvl3_5} and GPT5-Flash \cite{gpt5}. We utilized specific prompting strategies to guide the MLLMs to generate structured CoT reasoning alongside three-dimensional sub-scores regarding the visual quality, instruction alignment and content preservation, as well as an overall quality score. Finally, 113,898 CoT texts were successfully obtained.

\subsubsection{CoT scoring and evaluation}

\begin{figure}
    \centering
    \includegraphics[width=0.85\linewidth]{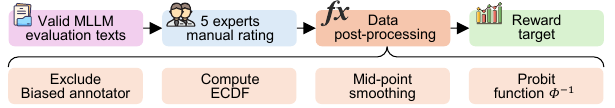}
    \caption{Annotation workflow}
    \label{fig:anno_flow}
\end{figure}

We then conducted the annotation for the CoT texts above via a custom-built web interface. Following the pipeline shown in Figure \ref{fig:anno_flow}, Each of the 113,898 valid MLLM evaluation samples were rigorously annotated by five trained expert raters. The raters evaluated each CoT text across the three aforementioned dimensions (logicality, accuracy, and usefulness) using a 5-point discrete Likert scale.

To ensure annotation fidelity, we applied a strict quality control and post-processing protocol. We first identified and excluded the data from one annotator who exhibited a severe structural bias (disproportionately assigning high scores with negligible variance). For the remaining annotations, we computed the Intra-class Correlation Coefficients  to quantify inter-rater reliability, achieving strong rank agreement (Kendall's W: 0.69-0.80) and moderate-to-good absolute agreement (ICC: 0.56-0.69) across the three dimensions. 

 The raw score distribution exhibits non-normality (Appendix \ref{sec:supp-postprocessing}). As this violates the normality assumption required for standard Z-score normalization, we employed a robust transformation pipeline:


First, we compute the Empirical Cumulative Distribution Function (ECDF) for each annotator $i$. Let $N$ denote the total number of samples evaluated by annotator $i$, and $C(x)$ be the count of samples receiving a score $\le x$. The raw percentile $P_i(x)$ is defined as Equation \ref{eq:raw_ecdf}. Next, to correct for the discrete jumps inherent in ordinal labels, we apply a mid-point smoothing adjustment (assuming $P_i(0) = 0$), as shown in Equation \ref{eq:smooth_ecdf}. Finally, we average the smoothed percentiles across all evaluating annotators to obtain an aggregated percentile $\bar{P} \in (0, 1)$. To provide a normally distributed continuous target suitable for downstream regression optimization, we apply the probit function (the inverse standard normal CDF, $\Phi^{-1}$) as Equation \ref{eq:probit_mapping}.

\begin{minipage}{0.21\textwidth}
    \begin{equation}
        P_i(x) = \frac{C(x)}{N}
        \label{eq:raw_ecdf}
    \end{equation}
\end{minipage}
\begin{minipage}{0.48\textwidth}
    \begin{equation}
        P'_i(x) = P_i(x - 1) + \frac{P_i(x) - P_i(x - 1)}{2}
        \label{eq:smooth_ecdf}
    \end{equation}
\end{minipage}
\begin{minipage}{0.26\textwidth}
    \begin{equation}
        \text{Reward} = \Phi^{-1}(\bar{P})
        \label{eq:probit_mapping}
    \end{equation}
\end{minipage}

The transformation pipeline above is theoretically robust and slightly superior regarding Leave-one-out PLCC and SRCC statistical test compared to utilizing either raw aggregated scores or standard Z-score normalization. Finally, we obtain $113,898\times4\times3=1,366,776$ valid human scores. These obtained high-confidence, unbiased human preference scores serve as the ground truth for training our CoT reward model. 

\subsubsection{High-quality CoT curation}

Regarding the construction of ReasonEdit-CoT-12K, we first filtered critics from ReasonEdit-Reward-113K to increase the diversity in language and avoid model-bias. Following the annotation, the filtering criteria contains: (1) The human-labelled scores on every dimension surpass 0.7; (2) The generated scores evaluating the source-edited pairs have the deviation from the EBench-18K human MOS is within 0.3 on every dimension. The second dataset yeilds 8K critics from 6 different MLLMs including Qwen, Intern and Gemini.

Given the limited ability of small-scale open-source MLLMs, we introduced SOTA large-scale MLLM (e.g. ChatGPT-5.4) to generate additional high-quality CoT texts. 50 hand-written high-quality CoT texts were randomly sampled as few-shot prompts. Finally, 5K CoT texts were selected, after filtering out those large score-deviation samples and human review.

Combining the two sources of critics mentioned above, after structural mismatch filtering and comprehensive human review by experts in image editing and evaluation, we yield a balanced, high-fidelity dataset of 12K CoT samples from 7 MLLMs, covering all source images and 67.13\% of source–edited pairs. This dataset also contains human MOS from EBench-18K.

\section{RE-Reward}
\label{para:reward_model}

\begin{figure}[htbp]
  \centering
  \includegraphics[width=\linewidth]{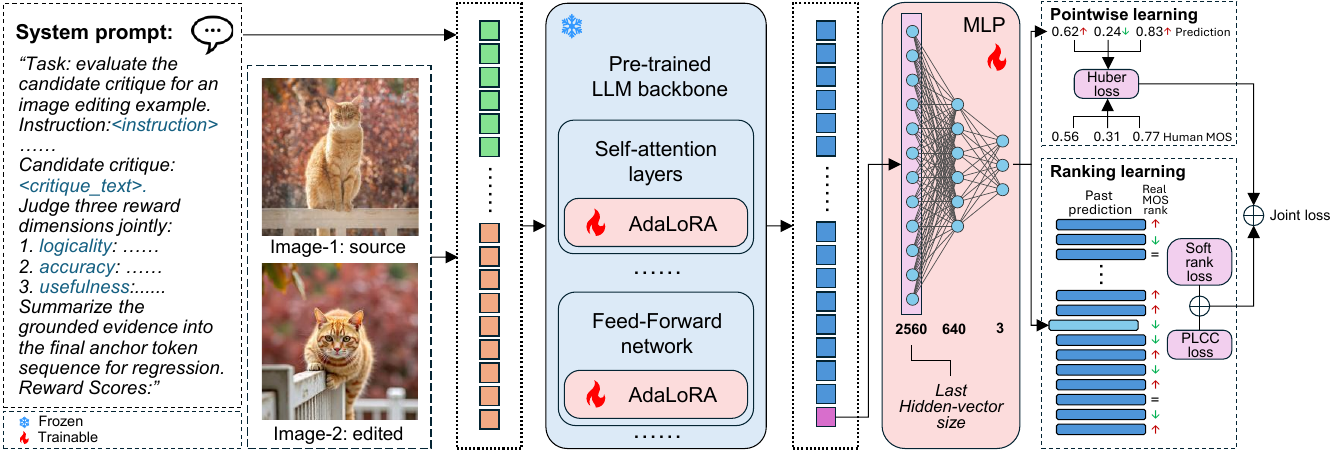}
  \caption{Overview of the RE-Reward architecture and SFT training}
  \label{fig:rm_model_arch} 
\end{figure}

In this section, we introduce \textbf{RE-Reward}, the first reward model designed for the explainable evaluation of image editing. While standard format rewards ensure structural compliance of the CoT reasoning, assessing the intrinsic semantic quality of the generated critique necessitates a fine-grained evaluator. To this end, we train a specialized multimodal reward model to automatically quantify the critique quality across three continuous dimensions: \textit{Logicality}, \textit{Accuracy}, and \textit{Usefulness}. The overall model architecture is shown in Figure \ref{fig:rm_model_arch}. 

We constructed this reward model using the Qwen3.5-4B backbone, utilizing Adaptive Low-Rank Adaptation (AdaLoRA) for parameter-efficient tuning, applying on all self-attention and MLP layers. To predict continuous scalar scores, we pool the last prompt token's final layer hidden state (prefill output) and append a customized multi-head Multi-Layer Perceptron (MLP) regression network. 

To capture both absolute scaling and relative preferences, we optimize the reward model using a composite objective:
\begin{equation}
    \mathcal{L}_{\text{RM}} = \lambda_{\text{huber}} \mathcal{L}_{\text{huber}} + \lambda_{\text{rank}} \mathcal{L}_{\text{rank}} + \lambda_{\text{plcc}} \mathcal{L}_{\text{plcc}}
\end{equation}
where $\mathcal{L}_{\text{huber}}$ provides robust absolute score regression resilient to annotation outliers. We also implemented a pairwise ranking loss $\mathcal{L}_{\text{rank}}$ across a massive pool of historical samples using a margin-based Softplus penalty. Furthermore, $\mathcal{L}_{\text{plcc}}$ maximizes the Pearson Linear Correlation Coefficient to ensure global distribution alignment between predictions and human annotations. 

In terms of training, given the massive number of samples derived from a relatively small set of original images (compared to the 113K evaluation samples), random sampling in standard training would lead to the frequent exposure of the same original image, introducing a high risk of overfitting. We employed a dynamic stratified sampling mechanism during training. (Detail in Appendix \ref{sec:supp_reward}).



\section{ReasonEdit}
In this section, we introduce \textbf{ReasonEdit}, focusing on interpretable evaluation of TIE. To address the limitations of existing black-box metrics, ReasonEdit takes a source image, an edited image, and a text instruction as inputs to provide a comprehensive assessment. It is designed not only to output continuous scalar scores for absolute quality regression, but also to generate highly logical, accurate, and useful Chain-of-Thought reasoning texts across three fine-grained dimensions: visual quality, instruction alignment, and content preservation.

\subsection{Model architecture}

\begin{figure}
  \centering
  \includegraphics[width=\linewidth]{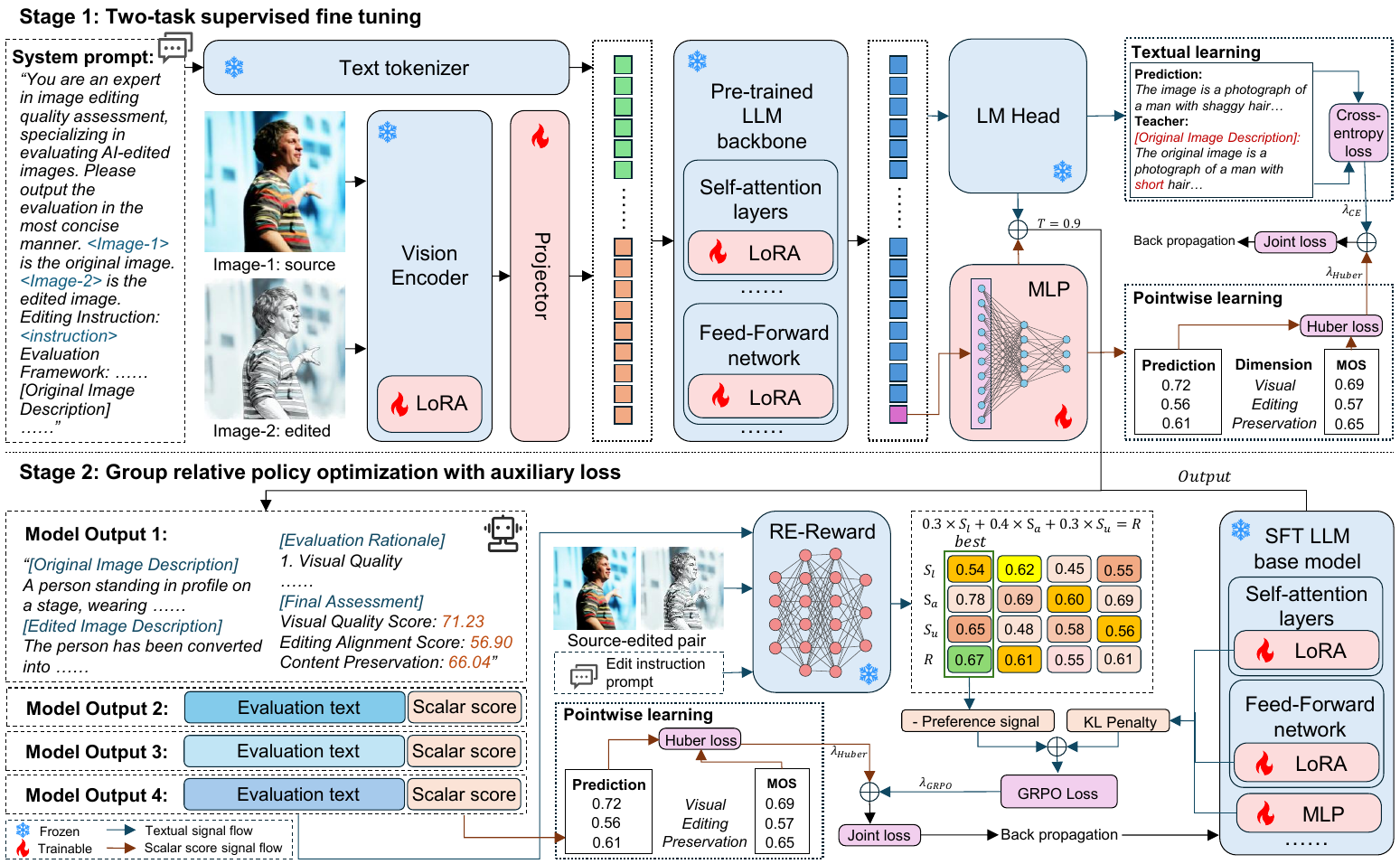}
  \caption{Overview of the ReasonEdit architecture and SFT training}
  \label{fig:model_arch} 
\end{figure}

The overall model architecture is shown in Figure \ref{fig:model_arch}. The input to our model is formulated as a multimodal triplet: the source image, the edited image, and the text prompt. The text prompt incorporates editing instructions and formatting directives. We adopt Qwen-3.5-9B\citep{qwen35} model as our MLLM backbone.

To achieve both CoT reasoning generation and 3-dimension continuous score prediction within a unified forward pass, our output architecture bifurcates into two distinct mechanisms. The first part of the output preserves the standard autoregressive text generation logic of the Qwen backbone, enabling the generation of CoT reasoning texts. Concurrently, an MLP regression head is attached to the backbone network to capture the prefill output and finally output three continuous scalar scores, each predicting the score for visual quality, instruction alignment, and content preservation. This hybrid architecture ensures that ReasonEdit successfully encapsulates both rich CoT text generation capabilities and fine-grained, continuous score prediction.

\subsection{Supervised fine-tuning process}

We conducted supervised fine-tunning (SFT) as the first-stage  (Figure \ref{fig:model_arch}, stage 1). The model was trained on ReasonEdit-CoT-12K dataset, which provides high-fidelity supervision signals comprising both valuable CoT texts and human-annotated MOS form EBench-18K\citep{lmm4edit}. We also used 4,000 additional samples with only MOS scores from EBench-18K\citep{lmm4edit} as regression signals for scalar scoring task. We utilized 4-bit Quantized Low-Rank Adaptive (QLoRA)\cite{QLoRA}, applying on all visual tower modules, self-attention layers and MLP layers. This strategy significantly enhances the model's representative capacity and human preference alignment.

The optimization is driven by a joint loss function that simultaneously supervises the text generation and score regression tasks. Specifically, the total loss is defined as a weighted sum of the standard autoregressive Cross-Entropy (CE) loss for the CoT text generation and the Huber loss for the continuous MOS regression. The joint training objective can be formulated as Equation \ref{eq:joint_loss}:

\begin{equation}
    \mathcal{L}_{\text{total}} = \lambda_{\text{CE}} \mathcal{L}_{\text{CE}} + \lambda_{\text{Huber}} \mathcal{L}_{\text{Huber}}
    \label{eq:joint_loss}
\end{equation}

$\lambda_{\text{CE}}$ and $\lambda_{\text{Huber}}$ are hyperparameter weights that balance the gradients between the semantic reasoning and the continuous scoring objectives. In our training, $\lambda_{\text{Huber}}=10$ while $\lambda_{\text{CE}}=1$.

We employ a multi-stage mixed training strategy, introducing 4,000 additional score-only samples from EBench-18K\citep{lmm4edit} in the intermediate phase to enhance data diversity and model generalization. The detailed epoch scheduling is available in the Appendix \ref{sec:supp_reasonedit}.

This scheduling strategy ensures a stable warm-up for both tasks initially. By introducing score-only samples in the middle phase, we improve the model's generalization capability through data diversity. Finally, reverting to dual-task training in the last phase prevents the model's weight distribution from being biased by the score-only samples.

\subsection{Reinforcement learning via GRPO}
While the initial policy model exhibits strong capabilities in CoT reasoning and format adherence, its assessment of specific image editing tasks tends to be overly harsh and misaligned with human perception. Armed with the reward model described in Section \ref{para:reward_model}, we conducted Reinforcement Learning from Human Feedback (RLHF) on the Supervised Fine-Tuned base model using Group Relative Policy Optimization (GRPO, Figure \ref{fig:model_arch}, stage 2)\citep{deepseekmath}.

The reward combines the scores $s_{log}$, $s_{acc}$ and $s_{use}$ from the three reward model outputs, \textit{\underline{log}icality score}, \textit{\underline{acc}uracy score} and \textit{\underline{use}fulness score}. Specifically, given the exact same triplet prompt, including the source image, edited image and instructions, we sample a group of $G=4$ distinct output responses from the old policy model. For each generated response, we assign a terminal reward formulated as follows:
\begin{equation}
\mathcal{R}_{RM}=0.3\,s_{log}+0.4\,s_{acc}+0.3\,s_{use}
\end{equation}
We assign a higher weight to accuracy, as factual correctness is the fundamental prerequisite for logicality and usefulness. Furthermore, we omit format-following and score-regression rewards, as the SFT model already handles formatting effectively, and the regression outputs are deterministic for a given prompt. Therefore, incorporating this metric within the context of GRPO would be meaningless.

To prevent the unconstrained RLHF process from degrading the regression head's performance, we introduce an auxiliary Huber loss to stabilize the dual-head model:
\begin{equation}
\mathcal{L} = \lambda_{grpo}\,\mathcal{L}_{GRPO} + \lambda_{mos}\,\mathcal{L}_{Huber}
\end{equation}
Here, $\mathcal{L}_{\mathrm{GRPO}}$ represents the negative sum of the Surrogate Objective Component and the KL Penalty Component from the standard GRPO formulation, while $\mathcal{L}_{\mathrm{Huber}}$ ensures the numerical stability of the regression head. The score regression capabilities are not compromised when applying GRPO.

The GRPO triplet prompts was sampled form ReasonEdit-22K constructed in Section \ref{data:all}. The samples with human MOS (from EBench-18K\citep{lmm4edit}) and pure source-edit pairs (new added) are sampled at a ratio of 7:3. 4,000 different triplet prompts were used in our GRPO training.
\begin{table}[t]
  \centering
  \caption{Performance comparisons of reward methods on ReasonEdit-Reward-113K from perspectives of \textit{logicality}, \textit{accuracy}, and \textit{usefulness}. SRCC ($\rho_s$), KRCC ($\rho_k$), and PLCC ($\rho_p$) metrics are reported. $\spadesuit$ Vision-language metrics, $\clubsuit$ small-scale MLLMs, \ding{72} middle-scale MLLMs, \ding{73} closed-source MLLMs. The best results are highlighted in \textbf{bold}, and the second-best results are \underline{underlined}.}
  \label{tab:reward_eval}
    \setlength{\tabcolsep}{10pt}
  \footnotesize
    \resizebox{1\textwidth}{!}{\begin{tabular}{ l ||ccc ccc ccc }
    \toprule
            \noalign{\vspace{-1.5pt}}
     \textbf{Dimensions} & \multicolumn{3}{c}{\textbf{Logicality}} & \multicolumn{3}{c}{\textbf{Accuracy}} & \multicolumn{3}{c}{\textbf{Usefulness}} \\
             \noalign{\vspace{-2pt}}
    \cmidrule(lr){2-4} \cmidrule(lr){5-7} \cmidrule(lr){8-10}
            \noalign{\vspace{-3pt}}
    \textbf{Method/Metric} & $\rho_s$ & $\rho_k$ & $\rho_p$ &  $\rho_s$ & $\rho_k$ & $\rho_p$ & $\rho_s$ & $\rho_k$ & $\rho_p$ \\
            \noalign{\vspace{-1.5pt}}
    \midrule
            \noalign{\vspace{-2.5pt}}
    
    $\spadesuit$PAC-S++ \citep{sarto2023positive} & 0.0298 & 0.0209 & 0.0148 & 0.0050 & 0.0029 & 0.0062 & 0.0169 & 0.0112 & 0.0181 \\
    $\spadesuit$FLEUR \citep{lee2024fleur} & 0.3384 & 0.2496 & 0.3258 & 0.4853 & 0.3612 & 0.5055 & 0.5393 & 0.4064 & 0.5422 \\
    $\spadesuit$CLIP-S-ViT-B/32 \citep{clipscore} & 0.0340 & 0.0238 & 0.0417 & 0.0655 & 0.0452 & 0.0624 & 0.0729 & 0.0501 & 0.0677 \\
            \noalign{\vspace{-2pt}}
    \midrule
            \noalign{\vspace{-2.5pt}}
    
    $\clubsuit$mPLUG-Owl3-7B \citep{mplug} & 0.3473 & 0.2819 & 0.3197 & 0.3835 & 0.3163 & 0.3894 & 0.0023 & 0.0019 & 0.0030 \\ 
    $\clubsuit$MiniCPM-V-2.6-8B \citep{minicpm} & 0.2749 & 0.2285 & 0.2269 & 0.1548 & 0.1287 & 0.1538 & 0.0000 & 0.0000 & 0.0000\\
    $\clubsuit$Ovis2.5-9B \citep{ovis25} & 0.0314 & 0.0268 & 0.0419 & 0.0752 & 0.0725 & 0.0453 & 0.1118 & 0.0747 & 0.1698 \\
    $\clubsuit$InternVL3-8B \citep{internvl3} & 0.1697 & 0.1440 & 0.1480 & 0.0687 & 0.0553 & 0.0929 & 0.2841 & 0.2410 & 0.2346 \\
    $\clubsuit$InternVL3.5-8B \citep{internvl3_5}: & 0.0784 & 0.0612 & 0.1798 & 0.0497 & 0.0410 & 0.0430 & 0.0736 & 0.0614 & 0.0291 \\
    $\clubsuit$Qwen3-VL-8B \citep{qwen3} & 0.5386 & 0.4545 & 0.5913 & 0.3947 & 0.3266 & 0.4175 & 0.3487 & 0.2907 & 0.3617 \\
    $\clubsuit$Qwen-3.5-9B \citep{qwen35} & 0.4097 & 0.3463 & 0.4787 & 0.4279 & 0.3544 & 0.4320 & 0.3906 & 0.3258 & 0.4037 \\
            \noalign{\vspace{-2pt}}
    \midrule
            \noalign{\vspace{-2.5pt}}
    \ding{72}DeepSeek-VL2-Small \citep{deepseekvl2} & 0.3318 & 0.2593 & 0.2843 & 0.4462 & 0.3525 & 0.3839 & 0.1848 & 0.1531 & 0.1881 \\
    \ding{72}Qwen-3.5-35B-A3B \citep{qwen35} & 0.5728 & 0.4860 & 0.5956 & 0.7761 & 0.6598 & 0.7797 & 0.7884 & 0.6729 & 0.7916 \\
    \ding{72}Gemma-4-31B \citep{gemma2024} & 0.7035 & 0.5902 & 0.7177 & 0.8264 & 0.6943 & 0.8251 & 0.8371 & 0.7078 & 0.8417 \\
            \noalign{\vspace{-2pt}}
    \midrule
            \noalign{\vspace{-2.5pt}}
    \ding{73}Gemini-3-Flash \citep{gemini3} & \underline{0.8438} & \underline{0.7136} & \underline{0.8173} & \underline{0.8587} & 0.7209 & 0.8411 & \underline{0.8676} & 0.7364 & 0.8609 \\
    \ding{73}ChatGPT-5.4 \citep{gpt5} & 0.7539 & 0.6528 & 0.7572 & 0.8532 & \underline{0.7284} & \underline{0.8535} & 0.8660 & \underline{0.7467} & \underline{0.8657}\\
            \noalign{\vspace{-2pt}}
    \midrule
            \noalign{\vspace{-2.5pt}}
              \rowcolor{gray!20}
    \textbf{RE-Reward (Ours)} & \textbf{0.8958} & \textbf{0.7315} & \textbf{0.9027} & \textbf{0.9323} & \textbf{0.7781} & \textbf{0.9316} & \textbf{0.9428} & \textbf{0.8015} & \textbf{0.9439}\\
                \noalign{\vspace{-1.5pt}}
    \bottomrule
  \end{tabular}}
\end{table}
\section{Experiments}
\subsection{Experiment setup}
All models were fine-tuned using LoRA/AdaLoRA on NVIDIA GPUs. Detailed hyperparameter configurations, learning rate schedules, and hardware setups are provided in the Appendix.
\subsection{RE-Reward performance}
To validate our reward model's alignment with human preferences, we compare its correlation with human annotations against diverse baselines—including vision-language metrics and state-of-the-art MLLMs—on the ReasonEdit-Reward-113K test set. For zero-shot evaluation, baseline MLLMs are prompted using a 5-point Likert scale to perform CoT reasoning before outputting a final JSON-formatted score.
\begin{table}
  \centering
  \caption{Performance comparisons of quality evaluation methods on EBench-18K from perspectives of visual quality, editing alignment, and content preservation. SRCC ($\rho_s$), KRCC-$\tau_b$ ($\rho_k$), and PLCC ($\rho_p$) metrics are reported. $\spadesuit$ Traditional FR IQA metrics, $\heartsuit$ traditional NR IQA metrics, $\clubsuit$ deep learning-based FR IQA methods, $\diamondsuit$ deep learning-based NR IQA methods, \ding{72} vision-language metrics, \ding{73} MLLM-based models. The best results are highlighted in \textbf{bold}, and the second-best results are \underline{underlined}.}
  \label{tab:performance_comparison}
  \footnotesize
  \setlength{\tabcolsep}{10pt}
  \resizebox{1\textwidth}{!}{\begin{tabular}{l ||ccc ccc ccc}
    \toprule
                \noalign{\vspace{-1.5pt}}
    \textbf{Dimensions} & \multicolumn{3}{c}{\textbf{Visual quality}} & \multicolumn{3}{c}{\textbf{Instruction alignment}} & \multicolumn{3}{c}{\textbf{Content preservation}} \\
                 \noalign{\vspace{-2pt}}
    \cmidrule(lr){2-4} \cmidrule(lr){5-7} \cmidrule(lr){8-10}
                \noalign{\vspace{-3pt}}
    \textbf{Methods/Metrics} & $\rho_s$ & $\rho_k$ & $\rho_p$ &  $\rho_s$ & $\rho_k$ & $\rho_p$ & $\rho_s$ & $\rho_k$ & $\rho_p$ \\
            \noalign{\vspace{-1.5pt}}
    \midrule
            \noalign{\vspace{-2.5pt}}
    
    $\spadesuit$SSIM\cite{SSIM} & 0.0035 & 0.0007 & 0.2133 & 0.1705 & 0.1132 & 0.2301 & 0.4635 & 0.3217 & 0.4865 \\
    $\spadesuit$SCSSIM\cite{SCSSIM} & 0.0640 & 0.0433 & 0.2646 & 0.2030 & 0.1350 & 0.2670 & 0.5868 & 0.4149 & 0.5938 \\
    $\spadesuit$GMSD\cite{GMSD} & 0.0099 & 0.0063 & 0.0959 & 0.2208 & 0.1485 & 0.2693 & 0.5272 & 0.3689 & 0.5328 \\
            \noalign{\vspace{-1.5pt}}
    \midrule
            \noalign{\vspace{-2.5pt}}
    $\heartsuit$DIIVINE\cite{DIIVINE} &0.1429&0.0929&0.3555&0.0471&0.0305&0.1196 &0.0121&0.0083&0.2051  \\
    $\heartsuit$BRISQUE\cite{BRISQUE} & 0.3423&0.2360&0.3955&0.1366&0.0923&0.1562&0.1302&0.0883&0.2095\\
    $\heartsuit$NIQE\cite{NIQE} &0.2979&0.2069&0.2453&0.1121&0.0763&0.1080&0.1748&0.1200&0.1926  \\
            \noalign{\vspace{-1.5pt}}
    \midrule
            \noalign{\vspace{-2.5pt}}
    $\clubsuit$LPIPS (alex) \cite{LPIPS} & 0.1832&0.1234&0.2782&0.2222&0.1489&0.2992&0.7395&0.5478&0.7594 \\
    $\clubsuit$LPIPS (vgg) \cite{LPIPS} & 0.1643&0.1101&0.1902&0.2166&0.1452&0.2632&0.7248&0.5326&0.7430 \\
    $\clubsuit$CVRKD$^*$\cite{CVRKD} &  0.7935&0.5991&0.8106&0.4661&0.3170&0.4806&0.7864&0.5917&0.8081\\
    $\clubsuit$AHIQ$^*$ \cite{AHIQ} & 0.8183&0.6241&0.8324&0.5249&0.3679&0.5452& 0.8365&0.6457&0.8515 \\
            \noalign{\vspace{-1.5pt}}
    \midrule
            \noalign{\vspace{-2.5pt}}
    $\diamondsuit$MANIQA$^*$ \cite{MANIQA} &0.8050 &0.6136 &0.8171 &0.3432&0.2661&0.3765 &0.6529&0.4716&0.7041\\
    $\diamondsuit$TOPIQ$^*$ \cite{TOPIQ}& 0.7936&0.6021&0.8054&0.3641&0.2536&0.3848 &0.6320&0.6692&0.4565\\
    $\diamondsuit$Q-Align$^*$ \cite{qalign}&0.8180&0.6285&0.8014&0.4961&0.3684&0.4994& 0.7046&0.5188&0.7321\\
            \noalign{\vspace{-1.5pt}}
    \midrule
            \noalign{\vspace{-2.5pt}}
    \ding{72}CLIPScore \cite{clipscore} & 0.2181 & 0.1467 & 0.2243 & 0.2152 & 0.1449 & 0.2209 & 0.2325 & 0.1586 & 0.2581 \\
    \ding{72}ImageReward \cite{imagereward} & 0.3991 & 0.2764 & 0.4351 & 0.2875 & 0.1978 & 0.3198 & 0.4033 & 0.2779 & 0.4662 \\
    \ding{72}PickScore \cite{pickscore}& 0.2483 & 0.1666 & 0.2889 & 0.3627 & 0.2482 & 0.3297 & 0.1357 & 0.0874 & 0.2046 \\
    \ding{72}VQAScore \cite{vqa}& 0.3014 & 0.2050 & 0.3162 & 0.2839 & 0.1898 & 0.2695& 0.2185 & 0.1444 & 0.2537  \\
            \noalign{\vspace{-1.5pt}}
    \midrule
            \noalign{\vspace{-2.5pt}}
    \ding{73}Qwen2-VL (7B) \cite{qwenvl} & 0.6786 & 0.4866 & 0.7041 & 0.1937 & 0.1427 & 0.1720 & 0.5478 & 0.3981 & 0.5200\\ 
    \ding{73}DeepSeekVL2 (small)$^*$ \cite{deepseekvl2} & 0.8674 & 0.7303 & 0.8665 & 0.8280 & 0.6893 & 0.8390 & 0.8778 & 0.7500 & 0.8819 \\
    \ding{73}InternVL2.5 (8B)$^*$ \cite{internvl2}& 0.8836 & 0.7223 & 0.8861 & 0.8207 & 0.6809 & 0.8387 & 0.8841 & 0.7571 & 0.8949 \\
    \ding{73}LMM4Edit $^*$ \cite{lmm4edit} & \underline{0.9136} & \underline{0.7432} &  \underline{0.9189} & \underline{0.8830} & \underline{0.7080} & \underline{0.8898} & \underline{0.9048} & \underline{0.7837} & \underline{0.9176} \\
            \noalign{\vspace{-1.5pt}}
    \midrule
            \noalign{\vspace{-2.5pt}}
          \rowcolor{gray!20}
    \textbf{ReasonEdit (Ours)} & \textbf{0.9272} & \textbf{0.7650} & \textbf{0.9311} & \textbf{0.9209} & \textbf{0.7552} & \textbf{0.9291} & \textbf{0.9466} & \textbf{0.8018} & \textbf{0.9524} \\
          \rowcolor{gray!20}
    Improvement (\%) & +1.49 & +2.93 & +1.33 & +4.29 & +6.67 & +4.42 & +4.62 & +2.31 & +3.79 \\
                    \noalign{\vspace{-1.5pt}}
    \bottomrule
  \end{tabular}}
\end{table}
\begin{table}[t]
  \centering
    \setlength{\tabcolsep}{12pt}
  \caption{Out-of-distribution test on other public image editing benchmarks. The best results are highlighted in \textbf{bold}, and the second-best results are \underline{underlined}. ``-'' indicates that the model was trained on this dataset, and thus its result is not reported to ensure a fair cross-dataset evaluation.}
  \label{tab:other_dataset}
  \footnotesize
  \resizebox{1\textwidth}{!}{\begin{tabular}{l||ccccc}
    \toprule
                \noalign{\vspace{-1.5pt}}
    \textbf{Methods/Benchmarks} & \makecell{\textbf{GenAI-Bench}\\\citep{genai} (Acc., \%)} & \makecell{\textbf{AROURA-Bench}\\\citep{aurora} (Acc., \%)} & \makecell{\textbf{EditScore-Bench}\\\citep{editscore} (Acc., \%)} & \makecell{\textbf{ImagenHub}\\\citep{imagenhub} (SRCC)} & \makecell{\textbf{IEQA}\\\citep{xu2026edithf1mmillionscalerichhuman} (SRCC)}\\
            \noalign{\vspace{-1.5pt}}
    \midrule
            \noalign{\vspace{-2.5pt}}
    GPT-4o \citep{chatgpt4o} & 53.54 & 50.81 & 67.30 & 0.3821 &0.3628\\
    GPT-5 \citep{gpt5} & 59.61 & 47.27 & \underline{75.24} & 0.4085 &0.4175\\
    Gemini-2.0-Flash \citep{gemini2} & 53.32 & 44.31 & 69.58 & 0.2369 &0.3512\\
    Gemini-2.5-Flash \citep{nanobanana} & 57.01 & 47.63 & 70.26 & \underline{0.4162} &0.4258\\
    Qwen3-VL (7B) \citep{qwen3} & 44.58 & 34.28 & 42.51 & 0.2426 &0.3860\\
    InternVL3.5 (7B) \citep{internvl3_5} & 45.17 & 36.92 & 45.18 & 0.2071 &0.3091\\
    EditScore (Qwen3) \citep{editscore} & 50.24 & 42.18 & 72.30 & 0.3062 &0.3805\\
    EditReward (MiMo) \citep{editreward} & \underline{65.72} & 63.62 & 71.25 & 0.3520 &0.4643\\
    LMM4Edit \citep{lmm4edit} & 63.27 & \underline{65.21} & 70.64 & 0.3726 &\underline{0.5012}\\
            \noalign{\vspace{-1.5pt}}
    \midrule
            \noalign{\vspace{-2.5pt}}
      \rowcolor{gray!20}
    \textbf{ReasonEdit (Ours)} & \textbf{83.90} & \textbf{72.22} & \textbf{78.48} & \textbf{0.7566} & \textbf{0.5830}\\
                \noalign{\vspace{-1.5pt}}
    \bottomrule
  \end{tabular}}
\end{table}
As reported in Table \ref{tab:reward_eval}, the evaluation is conducted using three standard statistical metrics: SRCC, KRCC ($\tau_b$) and PLCC. The results demonstrate that our fine-tuned 4B reward model consistently outperforms all MLLMs and traditional vision-language metrics. These findings confirm that our model is both highly accurate and computationally lightweight, making it an ideal reward signal for subsequent RL pipelines.
\subsection{ReasonEdit performance}
\paragraph{In-distribution test}In this experiment, we conduct a comparative evaluation on the in-distribution test set, EBench-18K\citep{lmm4edit}, to assess the correlation between the models' absolute scores and human preferences. The benchmark encompasses traditional IQA methods (both full-reference and no-reference), deep learning-based approaches, and LLM-based methods. As demonstrated in Table \ref{tab:performance_comparison}, our proposed RE-Reward achieves state-of-the-art (SOTA) performance across all three task categories and nine evaluation metrics, significantly outperforming the previous SOTA method, LLM4Edit\citep{lmm4edit}.

\paragraph{Out-of-distribution test}We further benchmarked our model on public zero-shot datasets, including GenAI-Bench\citep{genai}, AROURA-Bench\citep{aurora}, EditScore-Bench\citep{editscore}, ImagenHub\citep{imagenhub} and IEQA\citep{xu2026edithf1mmillionscalerichhuman}. For GenAI-Bench, AROURA-Bench and EditScore-Bench, we report the pairwise accuracy. For ImagenHub, we averaged our prediction on 3 dimensions as a overall score to match the annotations, SRCC was reported. For IEQA, in which all the scoring dimension could match, we calculated SRCC on 3 dimensions respectively and report the average SRCC. As demostrated in table \ref{tab:other_dataset}, our model achieves the highest performance across all zero-shot datasets, demonstrating state-of-the-art generalization capabilities.

\paragraph{CoT quality test}
In addition to absolute value regression, our model also demonstrates superior performance in generating explanatory CoT text. The table below presents a comparison of the quality of CoT generated by our model against that of leading open-source and closed-source models, benchmarked against the ground truth. On the test set of the EditReason-CoT-12k dataset, we employed metrics such as SimCSE \citep{SimCSE}, Word2Vec \citep{Word2Vec}, METEOR \citep{Meteor}, and ROUGE-1 \citep{ROUGE} for evaluation. The results, as shown in Table \ref{tab:quantitative_eval}, indicate that our model surpasses un-finetuned closed-source SOTA models, such as ChatGPT-5 and Gemini-2.5 Pro in both the quality of the generated text and the stability of its structure. 

\begin{table}
  \centering
  \caption{Quantitative CoT quality evaluation and ablation of ReasonEdit over five runs.}
  \label{tab:quantitative_eval}
  \footnotesize
  \setlength{\tabcolsep}{9pt}
  \renewcommand\arraystretch{1.2}
  \resizebox{1\textwidth}{!}{
  \begin{tabular}{l||ccccc}
    \toprule
                        \noalign{\vspace{-1.5pt}}
    \textbf{Method/Metric} & \textbf{MOS SRCC} & \textbf{SimCSE}\citep{SimCSE} & \textbf{Word2Vec}\citep{Word2Vec} & \textbf{Meteor}\citep{Meteor} & \textbf{ROUGE}\citep{ROUGE}\\
            \noalign{\vspace{-1.5pt}}
    \midrule
            \noalign{\vspace{-2.5pt}}
    GPT 5 Zero-Shot &
    0.4826 $\pm$ 0.0031 &
    0.6812 $\pm$ 0.0024 &
    0.6087 $\pm$ 0.0038 &
    0.1583 $\pm$ 0.0015 &
    0.1015 $\pm$ 0.0012 \\

    Gemini-2.5 Pro Zero-Shot &
    0.5024 $\pm$ 0.0027 &
    0.6765 $\pm$ 0.0021 &
    0.6121 $\pm$ 0.0032 &
    0.1610 $\pm$ 0.0013 &
    0.1042 $\pm$ 0.0010 \\
    
            \noalign{\vspace{-1.5pt}}
    \midrule
            \noalign{\vspace{-2.5pt}}

    Ovis2.5-9B \citep{ovis25} &
    0.4218 $\pm$ 0.0042 &
    0.6589 $\pm$ 0.0035 &
    0.6023 $\pm$ 0.0039 &
    0.1596 $\pm$ 0.0017 &
    0.0894 $\pm$ 0.0015 \\

    Ovis2.5-9B + GRPO &
    0.8382 $\pm$ 0.0061 &
    0.7015 $\pm$ 0.0044 &
    0.6158 $\pm$ 0.0050 &
    0.1712 $\pm$ 0.0021 &
    0.1149 $\pm$ 0.0019 \\

    Ovis2.5-9B + SFT &
    0.8826 $\pm$ 0.0048 &
    0.8123 $\pm$ 0.0037 &
    0.7510 $\pm$ 0.0045 &
    0.3385 $\pm$ 0.0026 &
    0.3327 $\pm$ 0.0031 \\

    \rowcolor{gray!20}
    \textbf{Ours} &
    \textbf{0.9079 $\pm$ 0.0035} &
    \textbf{0.8637 $\pm$ 0.0029} &
    \textbf{0.8442 $\pm$ 0.0032} &
    \textbf{0.3416 $\pm$ 0.0020} &
    \textbf{0.4518 $\pm$ 0.0028} \\

            \noalign{\vspace{-1.5pt}}
    \midrule
            \noalign{\vspace{-2.5pt}}

    InternVL3.5-8B \citep{internvl3_5} &
    0.4752 $\pm$ 0.0038 &
    0.6664 $\pm$ 0.0027 &
    0.5912 $\pm$ 0.0035 &
    0.1549 $\pm$ 0.0014 &
    0.0958 $\pm$ 0.0011 \\

    InternVL3.5-8B + GRPO &
    0.8221 $\pm$ 0.0057 &
    0.7096 $\pm$ 0.0042 &
    0.6427 $\pm$ 0.0048 &
    0.1765 $\pm$ 0.0022 &
    0.1420 $\pm$ 0.0017 \\

    InternVL3.5-8B + SFT &
    0.9013 $\pm$ 0.0041 &
    0.8554 $\pm$ 0.0030 &
    0.7859 $\pm$ 0.0039 &
    0.3412 $\pm$ 0.0025 &
    0.3075 $\pm$ 0.0029 \\

    \rowcolor{gray!20}
    \textbf{Ours} &
    \textbf{0.9215 $\pm$ 0.0032} &
    \textbf{0.8689 $\pm$ 0.0026} &
    \textbf{0.8483 $\pm$ 0.0031} &
    \textbf{0.3457 $\pm$ 0.0019} &
    \textbf{0.4296 $\pm$ 0.0025} \\

            \noalign{\vspace{-1.5pt}}
    \midrule
            \noalign{\vspace{-2.5pt}}

    Qwen3.5-VL-9B \citep{qwen35} &
    0.4960 $\pm$ 0.0034 &
    0.6517 $\pm$ 0.0029 &
    0.5978 $\pm$ 0.0036 &
    0.1526 $\pm$ 0.0013 &
    0.1708 $\pm$ 0.0016 \\

    Qwen3.5-VL-9B + GRPO &
    0.8435 $\pm$ 0.0055 &
    0.7893 $\pm$ 0.0041 &
    0.8354 $\pm$ 0.0046 &
    0.1637 $\pm$ 0.0018 &
    0.0941 $\pm$ 0.0012 \\

    Qwen3.5-VL-9B + SFT &
    0.9027 $\pm$ 0.0040 &
    0.8631 $\pm$ 0.0031 &
    0.8586 $\pm$ 0.0035 &
    0.3229 $\pm$ 0.0024 &
    0.4704 $\pm$ 0.0030 \\

    \rowcolor{gray!20}
    \textbf{Ours} &
    \textbf{0.9316 $\pm$ 0.0031} &
    \textbf{0.9164 $\pm$ 0.0025} &
    \textbf{0.9724 $\pm$ 0.0028} &
    \textbf{0.3508 $\pm$ 0.0018} &
    \textbf{0.5470 $\pm$ 0.0026} \\
                    \noalign{\vspace{-1.5pt}}
    \bottomrule
  \end{tabular}}
\end{table}

\subsection{Ablation study}
Table \ref{tab:quantitative_eval} presents our ablation study, where we compare standalone GRPO and SFT against our complete two-stage approach (SFT+GRPO) across different backbones, including Ovis2.5\citep{ovis25} and InternVL3.5\citep{internvl3_5}. We report the results and variance across five inference runs. First, applying GRPO directly to the base models proved highly effective, leading to a nearly twofold increase in SRCC. Employing SFT alone yielded even more substantial improvements, significantly enhancing both absolute score regression and the stable generation of high-quality CoT text. However, our combined SFT+GRPO strategy achieved the highest metrics across all models, demonstrating the efficacy of this dual-stage paradigm. Furthermore, these consistent gains across various backbones highlight our method's strong generalization capabilities and indirectly validate the high-quality, low-noise nature of our dataset.

\section{Conclusion}
In this paper, we introduce \textbf{ReasonEdit-22K}, a large-scale dataset for interpretable evaluation in TIE, featuring rich human feedbacks on evaluation texts and high-quality CoT samples. Built upon this dataset, we propose \textbf{RE-Reward}, a multimodal reward model that provides fine-grained, human-aligned feedback on evaluation text, and \textbf{ReasonEdit}, an interpretable evaluation framework trained with SFT and GRPO that learns to generate high-quality diagnostic evaluation text for image editing. Extensive experiments demonstrate that ReasonEdit achieves superior performance and generalizes well across public benchmarks, highlighting its potential for more transparent evaluation of TIE.



\medskip

\newpage
\clearpage
{
\small

\bibliographystyle{plain} 
\bibliography{references}
}

\newpage
\clearpage
\appendix
\clearpage
\begin{center}
\Large\textbf{Supplementary Material}
\end{center}

\newcommand{\suppblank}[1]{\textbf{[To be filled: #1]}}
\newcommand{\suppplaceholderfigure}[4][t]{
\begin{figure}[#1]
    \centering
    \fbox{\rule{0pt}{#2}\rule{0.92\linewidth}{0pt}}
    \caption{#3}
    \label{#4}
\end{figure}
}

\section{Overview}
\label{sec:supp_overview}
This supplementary material provides additional details for the data construction, annotation protocol, model design, training configuration, experiments, and responsible-release considerations of ReasonEdit. Section~\ref{sec:supp_data} describes the composition of ReasonEdit-22K, including the task taxonomy, source-edited pair organization, CoT generation, and human scoring protocol. Section~\ref{sec:supp_reward} details the architecture and training procedure of RE-Reward. Section~\ref{sec:supp_reasonedit} elaborates on the dual-head ReasonEdit evaluator, the supervised fine-tuning stage, and the GRPO stage. Section~\ref{sec:supp_comparison} gives additional information about baselines and benchmark protocols. Section~\ref{sec:supp_ablation} reserves additional ablation, uncertainty, and cost analyses. Finally, Sections~\ref{sec:supp_reproducibility}, \ref{sec:supp_limitations}, and \ref{sec:supp_impact} discuss reproducibility, limitations, licensing, ethics, and social impact.

\section{Details of ReasonEdit-22K}
\label{sec:supp_data}

\subsection{Dataset scope}
ReasonEdit-22K is designed for interpretable evaluation of text-guided image editing (TIE). Each core example is built around a triplet consisting of a source image, an edited image, and a textual editing instruction. The dataset supports two related forms of supervision. First, ReasonEdit-CoT-12K provides high-quality evaluation rationales and scalar editing-quality targets for training an interpretable evaluator. Second, ReasonEdit-Reward-113K provides multiple candidate CoT critiques and human ratings of critique quality for training RE-Reward.

The data construction follows the main-paper pipeline in Figure~\ref{fig:dataflow}: source-edited pairs are collected from EBench-18K \cite{lmm4edit} and from additional curated source images edited by recent TIE systems such as Qwen-Image-Edit \cite{qwenedit} and NanoBanana \cite{nanobanana}. The resulting candidate critiques are generated by multiple MLLMs and then rated by trained human annotators along logicality, accuracy, and usefulness.

\subsection{Task taxonomy}
The task taxonomy is organized to cover both high-level semantic editing and low-level restoration or enhancement. High-level edits evaluate whether the edited image correctly follows the semantic instruction while preserving unrelated content. Low-level edits evaluate whether visual degradations are removed or image fidelity is improved without introducing unintended semantic changes.

\textbf{High-level editing} includes the following instruction types:
\begin{itemize}[left=12pt, labelsep=0.6em, labelwidth=0pt]
\item \textbf{Add}: inserting a new object, attribute, or semantic element into the scene while maintaining scale, lighting, and context.
\item \textbf{Remove}: deleting a target object or region and plausibly completing the background.
\item \textbf{Replace}: substituting one object or entity with another while preserving scene layout and visual realism.
\item \textbf{Color}: changing the hue, tone, or color identity of a target object or region without changing its geometry.
\item \textbf{Texture}: modifying material appearance, surface detail, or local texture while preserving object identity.
\item \textbf{Style}: transferring the image into a new artistic, rendering, photographic, or visual style.
\item \textbf{Action}: changing a subject's action or state, which often requires compositional and physical reasoning.
\item \textbf{Weather and season}: altering environmental conditions such as rain, snow, sunshine, fog, or seasonal appearance.
\item \textbf{Expression}: modifying the facial expression of a human subject while preserving identity and facial structure.
\item \textbf{Background}: replacing or substantially changing the background while keeping the foreground target coherent.
\item \textbf{Counting}: changing the number of repeated objects or instances in a controlled way.
\item \textbf{Position}: moving a target object to a new location while maintaining perspective and occlusion relationships.
\item \textbf{Size}: enlarging or shrinking a target object without changing its semantic identity.
\item \textbf{Lighting}: changing illumination strength, direction, color temperature, or scene mood.
\item \textbf{Identity}: preserving or changing person-specific identity cues in portrait-centered edits.
\item \textbf{Outpainting}: extending the image canvas beyond the original field of view with coherent content.
\item \textbf{Crop and zoom}: changing framing or magnification to emphasize a target region.
\item \textbf{Composite multi-edit}: combining multiple editing operations in a single instruction.
\end{itemize}

\textbf{Low-level editing} includes the following restoration and enhancement types:
\begin{itemize}[left=12pt, labelsep=0.6em, labelwidth=0pt]
\item \textbf{Deblur}: recovering sharp structure from motion blur or defocus blur.
\item \textbf{Dehaze}: removing haze or fog-like veiling effects to recover visibility and contrast.
\item \textbf{Denoise}: suppressing sensor, compression, or synthetic noise while preserving details.
\item \textbf{Derain}: removing rain streaks, droplets, or rain-induced occlusions.
\item \textbf{Desnow}: clearing snow artifacts and restoring scene visibility.
\item \textbf{Low-light enhancement}: improving brightness and details in dark images without excessive noise amplification.
\item \textbf{Shadow removal}: reducing unwanted cast shadows while keeping natural lighting transitions.
\item \textbf{Super-resolution}: reconstructing higher-frequency details from low-resolution inputs.
\end{itemize}

\begin{figure}
    \centering
    \includegraphics[width=0.9\linewidth]{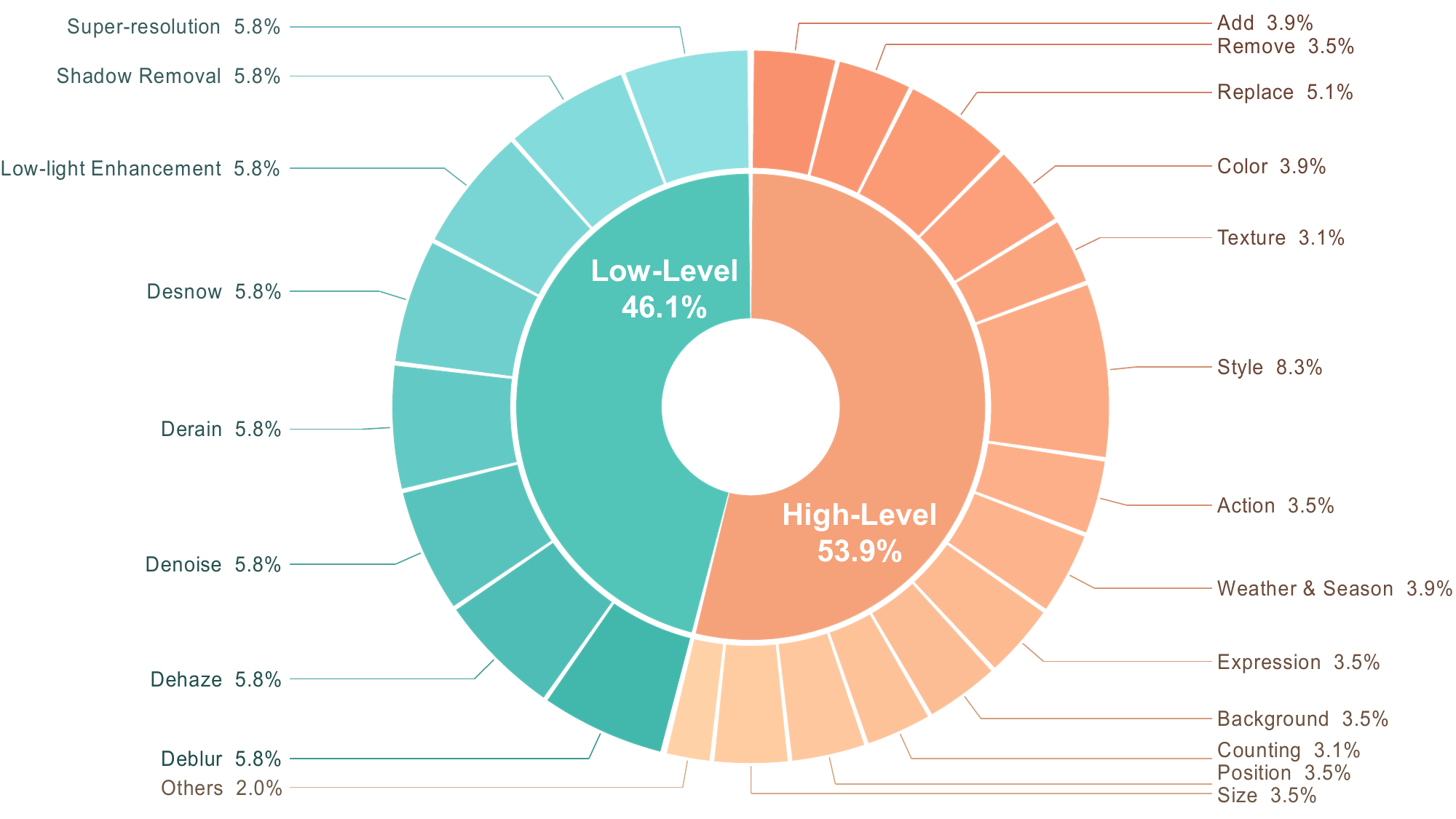}
    \caption{Task-type distribution in ReasonEdit-22K}
    \label{fig:datacat}
\end{figure}

\subsection{Candidate CoT generation}
\label{sec:supp_cot_generation}
For each triplet, candidate CoT critiques are generated by multiple MLLMs. The prompt requires the model to compare the source image and edited image under the editing instruction, discuss visual quality, instruction alignment, and content preservation, and then produce scalar assessment lines. The candidate-generator pool includes open-source and closed-source MLLMs used in the main paper, such as Qwen3-VL \cite{qwen3}, InternVL3.5 \cite{internvl3_5}, and GPT-5-family models \cite{gpt5}. Metadata are shown in \ref{tab:evaluator_models}.

  \begin{table}
  \small
  \centering
  \caption{Metadata of evaluator models used to generate critique texts in the manifest}
  \label{tab:evaluator_models}
  \begin{tabular}{llrrr}
  \toprule
  Model & Scale & Access & Critiques & Image pairs \\
  \midrule
  GPT-5 Mini\cite{gpt5} & undisclosed & closed-source & 12{,}672 & 12{,}672 \\
  Gemini 3 Flash\cite{gemini3} & undisclosed & closed-source & 12{,}660 & 12{,}660 \\
  InternVL3.5-8B\cite{internvl3_5} & 8B & open-source & 12{,}670 & 12{,}670 \\
  LLaVA-1.6-7B\cite{llava} & 7B & open-source & 12{,}653 & 12{,}653 \\
  Llama-3.2-Vision-11B\cite{llama} & 11B & open-source & 12{,}655 & 12{,}655 \\
  MiniCPM-V-2.6\cite{minicpm} & 8B & open-source & 12{,}580 & 12{,}580 \\
  Ovis2.5-9B\cite{ovis25} & 9B & open-source & 12{,}669 & 12{,}669 \\
  Qwen3-VL\cite{qwen3} & 8B & open-source & 12{,}666 & 12{,}666 \\
  mPLUG-Owl3-7B\cite{mplug} & 7B & open-source & 12{,}672 & 12{,}672 \\
  \bottomrule
  \end{tabular}
  \end{table}

\subsection{Human scoring annotation}
\label{sec:supp_annotation}
The scoring annotation evaluates the quality of a candidate critique rather than the edited image directly. Each annotator is shown the source image, the edited image, the editing instruction, and a candidate CoT critique. The annotator then assigns three 5-point Likert scores:
\begin{itemize}[left=12pt, labelsep=0.6em, labelwidth=0pt]
\item \textbf{Logicality}: whether the critique follows a coherent reasoning chain and reaches conclusions supported by its own observations.
\item \textbf{Accuracy}: whether the critique is faithful to the source image, edited image, and instruction without hallucinated or incorrect visual claims.
\item \textbf{Usefulness}: whether the critique provides actionable and diagnostic information for evaluation, comparison, or model improvement.
\end{itemize}

\begin{figure}[t]
    \centering
    \includegraphics[width=\linewidth]{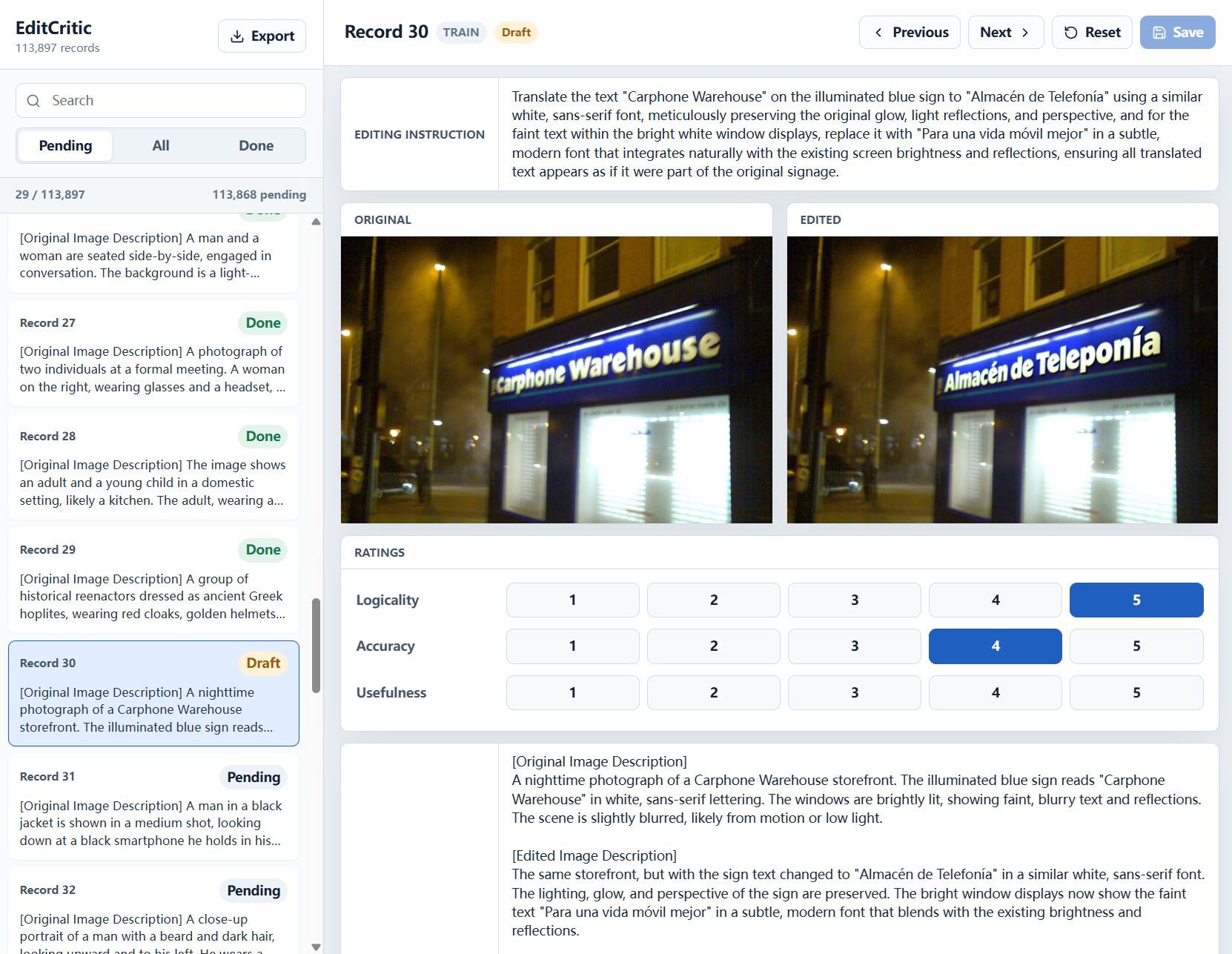}
    \caption{Annotation interface for scoring candidate critiques}
    \label{fig:supp_annotation_interface}
\end{figure}

\subsubsection{Annotator training and quality control}
Before formal annotation, annotators complete a calibration stage with examples covering successful edits, obvious failures, borderline cases, and critiques with hallucinated observations. During annotation, examples are assigned in randomized order. We monitor score variance, per-annotator score distribution, and inter-rater agreement to detect structural bias. The main paper reports that one annotator with severe structural bias was excluded before final score aggregation.

\subsubsection{Likert scoring rubric}
Table~\ref{tab:supp_likert_rubric} gives the detailed rubric used to align annotators. A high score requires not only fluent writing but also faithful visual grounding and diagnostic value.

\begin{table*}[t]
\centering
\caption{5-point Likert rubric for evaluating candidate CoT critiques.}
\label{tab:supp_likert_rubric}
\small
\setlength{\tabcolsep}{6pt}
\resizebox{\textwidth}{!}{
\begin{tabular}{p{0.9cm} p{5.0cm} p{5.0cm} p{5.0cm}}
\toprule
\textbf{Score} & \textbf{Logicality} & \textbf{Accuracy} & \textbf{Usefulness} \\
\midrule
1 & The critique is contradictory, disorganized, or unsupported by a coherent reasoning chain. & The critique contains major factual errors, hallucinations, or clear misreadings of the source image, edited image, or instruction. & The critique is too vague, misleading, or irrelevant to support diagnosis or reward-model training. \\
\midrule
2 & The critique has a partially understandable structure, but important reasoning links are missing, weak, or inconsistent. & The critique captures a few correct observations but includes substantial unsupported claims or factual mistakes. & The critique provides limited actionable information and misses the main evidence needed for evaluation. \\
\midrule
3 & The critique is broadly understandable and mostly ordered, but some reasoning steps remain shallow or only partially justified. & The critique is mostly correct, though it may omit important visual evidence or include minor questionable claims. & The critique is moderately helpful but lacks specificity or diagnostic depth in important places. \\
\midrule
4 & The critique is logically organized, internally consistent, and grounded in a clear evaluation flow. & The critique is largely faithful to the images and instruction, with only minor imprecision. & The critique is useful for diagnosing quality, identifying failure modes, and supporting model comparison. \\
\midrule
5 & The critique is fully coherent, precise, and persuasive, with conclusions explicitly supported by visual evidence. & The critique accurately reflects the instruction and both images without hallucinated details. & The critique is highly diagnostic, specific, and directly valuable for human interpretation, reward modeling, and model improvement. \\
\bottomrule
\end{tabular}}
\end{table*}

\subsubsection{Score post-processing}
\begin{figure}
    \centering
    \includegraphics[width=0.9\linewidth]{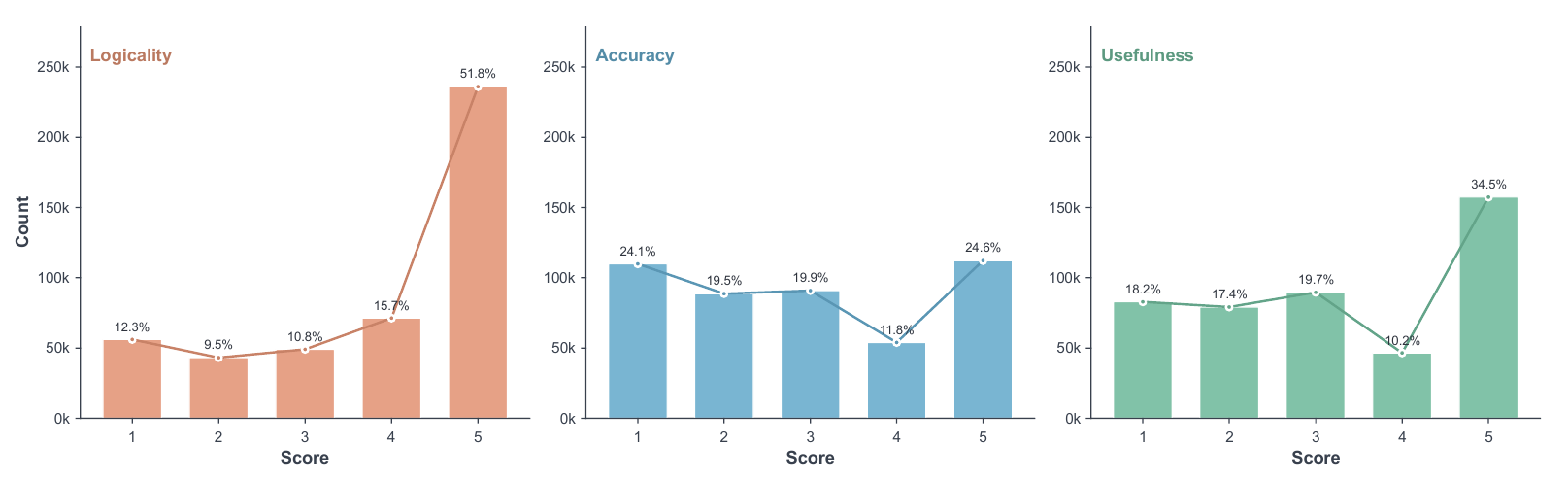}
    \caption{The raw 1–5 ordinal scores after the annotation of ReasonEdit-Reward-113K}
    \label{fig:raw_score}
\end{figure}
The raw 1--5 ordinal scores are discrete and non-Gaussian, as shown in Figure \ref{fig:raw_score}. Therefore, direct Z-score normalization is not ideal. Following the main paper, we use an annotator-wise empirical cumulative distribution function (ECDF), midpoint smoothing for ordinal bins, annotator averaging, and probit mapping. For annotator $i$, let $N_i$ be the number of scored samples and let $C_i(x)$ count samples receiving score $\le x$. The raw percentile is
\begin{equation}
    P_i(x)=\frac{C_i(x)}{N_i}.
    \label{eq:supp_raw_ecdf}
\end{equation}
The midpoint-smoothed percentile is
\begin{equation}
    P'_i(x)=P_i(x-1)+\frac{P_i(x)-P_i(x-1)}{2},
    \label{eq:supp_smooth_ecdf}
\end{equation}
where $P_i(0)=0$. For a candidate critique scored by multiple annotators, the smoothed percentiles are averaged into $\bar{P}$ and mapped to a continuous target by
\begin{equation}
    y=\Phi^{-1}(\bar{P}).
    \label{eq:supp_probit_mapping}
\end{equation}
This transformation preserves annotator ranking information while producing continuous targets suitable for regression-based reward-model training.

\subsection{Generation of teacher high-quality texts}
We employed flagship models, including the GPT-5 series \cite{gpt5} and Gemini-3 series \cite{gemini3}, to augment our dataset with high-quality text. The base prompt used for generation is as follows:
\begin{tcolorbox}[title=Prompt Template, colback=gray!5, colframe=black]
You are an expert in image editing quality assessment, specializing in evaluating AI-edited images. Please output the evaluation in the most concise manner. \\
Original Image: \verb+\img1+ \\
Edited Image: \verb+\img2+ \\
Editing Instruction: \verb+<instruction_text>+ \\
Evaluation Framework:\\
\textbf{\lbrack Original Image Description\rbrack}\\
What is the content of the original image?\\
\textbf{\lbrack Edited Image Description\rbrack}\\
What is the content of the edited image?\\
\textbf{\lbrack Evaluation Rationale\rbrack} \\
Evaluate based on the following three dimensions, providing at least two points for each.\\
1. Visual Quality (Naturalness of the edit and image)\\
E.g., lighting, clarity, color, details, realism, etc.\\
2. Editing Alignment (Adherence to editing instructions)\\
Whether the instruction is fully or partially implemented, and the effectiveness of the implementation.\\
3. Content Preservation (Content consistency)\\
E.g., consistency of the main structure with the original, preservation of unedited areas, style consistency.\\
\textbf{\lbrack Final Assessment\rbrack}\\
After outputting [Final Assessment], immediately continue with exactly three scores for Visual Quality, Editing Alignment, and Content Preservation in one line, separated by commas, with no extra words.\\
Use the format X.XX, X.XX, X.XX with exactly two decimal places.\\
Example: [Final Assessment]0.58, 0.36, 0.50
\end{tcolorbox}
If the prompt failed to yield satisfactory results—specifically, if the deviation between the model's scoring and human ratings on a specific dimension of the edited image's 3D attributes exceeded 0.15 (after normalization)—a resampling mechanism was automatically triggered. We injected specific instructions into the base prompt to guide the model to focus on previously underperforming aspects, such as:
\begin{tcolorbox}[title=Prompt Template, colback=gray!5, colframe=black]
Editing: You must pay special attention to whether the edited image fully follows the instruction. Any detail mismatch, such as incorrect color, wrong count, wrong position, or omission of a secondary description, should be treated as failure. Even if the image looks attractive, you must lower the score if the instruction is not followed completely.
\end{tcolorbox}
\begin{tcolorbox}[title=Prompt Template, colback=gray!5, colframe=black]
Visual Quality: Use professional photography and top-tier design imagery as the standard. Any noise, blur, lighting inconsistency, anatomical defect such as flawed fingers, object distortion, deformation, or unnatural texture must reduce the score. The image must reach high-definition and near-perfect quality.
\end{tcolorbox}
Ultimately, 8K samples passed the sampling phase and were forwarded to subsequent fine-tuning and manual selection pipelines. This process resulted in 5K high-quality CoT (Chain-of-Thought) texts.

\subsection{Human-subject and labor reporting}
A total of five annotators were involved in this annotation process, with four of them passing both the pre-annotation calibration and the quality acceptance checks. All five annotators provided informed consent to participate in and withdraw from the process, and received compensation based on working hours that exceeded the local minimum wage standard.

The four qualified annotators spent approximately 900 hours scoring evaluation texts and about 10 hours drafting zero-shot example samples for generating high-quality CoT. An additional 150 hours were dedicated to the acceptance and fine-tuning of the 12,000 high-quality CoT texts selected by the screening mechanism. Although the fifth annotator's scores were not included in the final calculation, we acknowledge his efforts and referenced his feedback during the high-quality CoT text acceptance and fine-tuning phases. No obvious physical injury risks were identified during this annotation process.

\section{Details of RE-Reward}
\label{sec:supp_reward}

\subsection{Architecture}
RE-Reward is a multimodal reward model for judging the quality of candidate evaluation critiques. It takes the source image, edited image, editing instruction, and candidate critique as input, then predicts three continuous scores corresponding to logicality, accuracy, and usefulness. The model is built on a Qwen3.5-4B multimodal backbone with AdaLoRA adapters. The regression head pools the final valid token representation and applies a lightweight MLP:
{\small
\[
\texttt{Linear(hidden\_size, hidden\_size/4)}
\rightarrow \texttt{SiLU}
\rightarrow \texttt{Dropout(0.15)}
\rightarrow \texttt{Linear(hidden\_size/4, 3)}
\]
}
The three outputs are trained against the post-processed human targets described in Section~\ref{sec:supp_annotation}.

\subsection{Training objective}
The reward-model objective combines absolute regression, pairwise ranking, and correlation alignment:
\begin{equation}
    \mathcal{L}_{\mathrm{RM}}
    =
    \lambda_{\mathrm{huber}}\mathcal{L}_{\mathrm{huber}}
    +
    \lambda_{\mathrm{rank}}\mathcal{L}_{\mathrm{rank}}
    +
    \lambda_{\mathrm{plcc}}\mathcal{L}_{\mathrm{plcc}}.
    \label{eq:supp_rm_loss}
\end{equation}
The Huber term stabilizes regression under noisy human labels. The ranking term encourages the reward model to preserve relative critique quality across samples. The PLCC term aligns the global linear correlation between predictions and human targets, which is important because RE-Reward later provides scalar rewards for GRPO.

\subsection{Dynamic stratified sampling}
ReasonEdit-Reward contains many candidate critiques for a smaller set of source-edited pairs. Naive random sampling could repeatedly expose the model to the same source image or edited pair, increasing source-level overfitting. The training code addresses this by grouping examples using image-pair identifiers and source-image identifiers, then limiting the number of pair groups sampled from the same source image per epoch. Validation and test splits use deterministic sampling to keep evaluation stable. Specifically, within each epoch, we applied stratified sampling to ensure that the original-edited image pairs derived from any single original image appeared no more than 6 times. Simultaneously, the evaluation text derived from each original-edited image pair was exposed a maximum of 3 times. Using the lack of improvement in validation set performance for two consecutive epochs as the early stopping criterion, the model was ultimately trained for 9 epochs. According to log statistics, all 113K samples were exposed at least once.

\subsection{Reward-model training configuration}
Table~\ref{tab:supp_rm_config} records the implementation-level configuration used by the checked training code.

\begin{table*}[htbp]
\centering
\caption{Implementation details of RE-Reward training. Update from the final saved training configuration before release.}
\label{tab:supp_rm_config}
\small
\setlength{\tabcolsep}{7pt}
\resizebox{\textwidth}{!}{
\begin{tabular}{p{4.0cm} p{12.0cm}}
\toprule
\textbf{Category} & \textbf{Setting} \\
\midrule
Backbone & Qwen3.5-4B multimodal model \\
\midrule
Adapter & AdaLoRA on attention and MLP projection modules \\
AdaLoRA prameters & initial rank $64$, target rank $32$, $\alpha=32$, dropout $0.05$ \\
\midrule
Precision and memory & \texttt{bf16}; gradient checkpointing enabled \\
\midrule
Sequence and image budget & maximum sequence length $1536$; image pixel cap 262,144 \\
\midrule
Batching & per-device batch size $1$; gradient accumulation $8$ \\
\midrule
Optimizer & AdamW; learning rate $1\times10^{-4}$; weight decay $0.01$; $(\beta_1,\beta_2,\epsilon)=(0.9,0.98,10^{-8})$ \\
\midrule
Schedule & linear warmup ratio $0.08$ followed by cosine decay to minimum LR ratio $0.3$ \\
\midrule
Training horizon & up to $9$ epochs with early stopping patience $2$ and minimum improvement $10^{-4}$ \\
\midrule
Loss weights & $\lambda_{\mathrm{huber}}=4.0$, $\lambda_{\mathrm{rank}}=0.05$, $\lambda_{\mathrm{plcc}}=0.05$ \\
\midrule
Auxiliary queues & rank queue size $4096$; rank-label margin $0.03$; PLCC queue size $256$ \\
\bottomrule
\end{tabular}}
\end{table*}

\subsection{Reward serving during GRPO}
During GRPO training, RE-Reward is served by a local HTTP service. For a generated critique, the server returns logicality, accuracy, usefulness, and the weighted scalar reward
\begin{equation}
    \mathcal{R}_{\mathrm{RM}}
    =
    0.3\,s_{\mathrm{log}}
    +
    0.4\,s_{\mathrm{acc}}
    +
    0.3\,s_{\mathrm{use}}.
    \label{eq:supp_weighted_reward}
\end{equation}
The accuracy dimension receives the largest weight because factual faithfulness to the images and instruction is a prerequisite for useful reasoning, and MLLMs inherently do better in other two aspects. The serving code uses a batched FastAPI endpoint and validates each request with source image, edited image, instruction, and critique fields.

\subsection{Reward-model prompt format}
The reward model receives a textual prompt that asks it to evaluate a candidate critique for an image-editing example. The prompt includes the editing instruction, the candidate critique, and a final ``Reward Scores'' cue:

\begin{tcolorbox}[title=Prompt Template, colback=gray!5, colframe=black]
Task: evaluate the candidate critique for an image editing example. Image-1 is the original source image and Image-2 is the edited result image.

Editing instruction:
\verb+<instruction_text>+

Candidate critique:
\verb+<critic_text>+

Judge three reward dimensions jointly: \\
1. logicality: internal consistency, coherent reasoning, and absence of contradictions. \\
2. accuracy: factual alignment with the source image, edited image, and editing instruction. \\
3. usefulness: specificity, diagnostic value, and usefulness for reward modeling. \\
Summarize the grounded evidence into the final anchor token sequence for regression. \\
Reward Scores:
\end{tcolorbox}

\section{Details of ReasonEdit}
\label{sec:supp_reasonedit}

\subsection{Dual-head model architecture}
ReasonEdit is a multimodal generator-regressor for interpretable TIE evaluation. It takes the source image, edited image, and editing instruction as input. The autoregressive language-model head generates a structured CoT critique, while a separate regression head predicts three scalar editing-quality scores for visual quality, instruction alignment, and content preservation.

The final normalized hidden state is pooled and passed through the regression MLP:
{\small
\[
\texttt{Linear(hidden\_size, hidden\_size/2)}
\rightarrow \texttt{SiLU}
\rightarrow \texttt{Dropout(0.1)}
\rightarrow \texttt{Linear(hidden\_size/2, 3)}
\]
}

\subsection{Supervised fine-tuning}
The SFT stage trains ReasonEdit on ReasonEdit-CoT-12K using both text-generation supervision and scalar regression supervision. Let $\mathcal{L}_{\mathrm{CE}}$ denote the autoregressive cross-entropy loss for the CoT target text, and let $\mathcal{L}_{\mathrm{Huber}}$ denote the regression loss for the three scalar scores. The joint objective is
\begin{equation}
    \mathcal{L}_{\mathrm{SFT}}
    =
    \lambda_{\mathrm{CE}}\mathcal{L}_{\mathrm{CE}}
    +
    \lambda_{\mathrm{reg}}\mathcal{L}_{\mathrm{Huber}}.
    \label{eq:supp_sft_loss}
\end{equation}
The main paper describes this stage as a mixed training process with high-quality CoT samples and additional score-only samples from EBench-18K \cite{lmm4edit}. The final schedule is shown in Table \ref{tab:supp_sft_schedule}.

\begin{table}
\centering
\caption{Final mixed SFT schedule}
\label{tab:supp_sft_schedule}
\small
\setlength{\tabcolsep}{8pt}
\begin{tabular}{p{3.2cm} p{2.0cm} p{3.5cm} p{3.0cm}}
\toprule
\textbf{Phase} & \textbf{Epoch range} & \textbf{Data composition} & \textbf{Active losses} \\
\midrule
Dual-task warm-up & 1 & 10K score+CoT & CE + Huber \\
\midrule
Mixed score-only stage & 2, 3, 4 & \makecell[l]{10K score+CoT\\4K score only} & \makecell[l]{CE(for samples\\with CoT) + Huber} \\
\midrule
Final dual-task refresh & 5, 6 & 10K score+CoT & CE + Huber \\
\bottomrule
\end{tabular}
\end{table}

\subsection{SFT configuration}
Table~\ref{tab:supp_sft_config} summarizes the implementation detail.

\begin{table*}
\centering
\caption{Implementation details of ReasonEdit SFT. Update from the final saved training configuration before release.}
\label{tab:supp_sft_config}
\small
\setlength{\tabcolsep}{7pt}
\resizebox{\textwidth}{!}{
\begin{tabular}{p{4.0cm} p{12.0cm}}
\toprule
\textbf{Category} & \textbf{Setting} \\
\midrule
Backbone & Qwen3.5-9B multimodal model \cite{qwen35} \\
\midrule
Adapter / quantization & 4-bit loading with LoRA adapters \\
\midrule
LoRA modules & \makecell[l]{"up\_proj",
    "gate\_proj",
    "attn.qkv",
    "patch\_embed.proj",
    "k\_proj",
    "q\_proj",\\
    "merger.linear\_fc2",
    "mlp.linear\_fc1",
    "down\_proj",
    "o\_proj",\\
    "mlp.linear\_fc2",
    "attn.proj",
    "merger.linear\_fc1",
    "v\_proj"}\\
\midrule
LoRA parameters & rank $64$, alpha $128$, dropout $0.05$ \\
\midrule
Sequence and image budget & maximum sequence length $3072$; maximum pixels per image $262,144$ \\
\midrule
Batching & per-device batch size $1$; gradient accumulation $4$ \\
\midrule
Optimizer & AdamW; LoRA learning rate $1\times10^{-5}$; regression-head learning rate $2\times10^{-5}$; weight decay $0.01$ \\
\midrule
Schedule & warmup ratio $0.03$ with cosine decay; validation at a configurable epoch fraction \\
\midrule
Losses & CE for generated critique text plus Huber regression for visual quality, instruction alignment, and content preservation \\
\bottomrule
\end{tabular}}
\end{table*}

\subsection{GRPO training}
The GRPO stage refines the SFT policy using RE-Reward as the terminal reward model. Given the same source image, edited image, and instruction, the policy samples a group of $G$ responses. RE-Reward scores each response, and GRPO normalizes rewards within each group to form relative advantages. The objective includes the standard policy optimization term with a reference-model KL penalty \cite{deepseekmath}. The main paper uses $G=4$ and the reward in Equation~\ref{eq:supp_weighted_reward}.

The implementation additionally supports auxiliary Huber supervision for the regression head on prompts with available MOS targets. This prevents RL updates that improve generated text from degrading scalar-score prediction. The trainer can also distinguish strong MOS supervision from weaker reference supervision when such metadata is available.

\subsection{GRPO configuration}
Please refer to Table \ref{tab:supp_grpo_config} for details.

\subsection{Inference protocol}
At inference time, ReasonEdit receives the same triplet input used during training.

\begin{tcolorbox}[title=Prompt Template, colback=gray!5, colframe=black]
You are an expert in image editing quality assessment, specializing in evaluating AI-edited images. Please output the evaluation in the most concise manner. \\
Original Image: \verb+\img1+ \\
Edited Image: \verb+\img2+ \\
Editing Instruction: \verb+<instruction_text>+ \\
Evaluation Framework:\\
\textbf{\lbrack Original Image Description\rbrack}\\
What is the content of the original image?\\
\textbf{\lbrack Edited Image Description\rbrack}\\
What is the content of the edited image?\\
\textbf{\lbrack Evaluation Rationale\rbrack} \\
Evaluate based on the following three dimensions, providing at least two points for each.\\
1. Visual Quality (Naturalness of the edit and image)\\
E.g., lighting, clarity, color, details, realism, etc.\\
2. Editing Alignment (Adherence to editing instructions)\\
Whether the instruction is fully or partially implemented, and the effectiveness of the implementation.\\
3. Content Preservation (Content consistency)\\
E.g., consistency of the main structure with the original, preservation of unedited areas, style consistency.\\
\textbf{\lbrack Final Assessment\rbrack}\\
After outputting [Final Assessment], immediately continue with exactly three scores for Visual Quality, Editing Alignment, and Content Preservation in one line, separated by commas, with no extra words.\\
Use the format X.XX, X.XX, X.XX with exactly two decimal places.\\
Example: [Final Assessment]0.58, 0.36, 0.50
\end{tcolorbox}

\begin{table*}
\centering
\caption{Implementation details of ReasonEdit GRPO}
\label{tab:supp_grpo_config}
\small
\setlength{\tabcolsep}{7pt}
\resizebox{\textwidth}{!}{
\begin{tabular}{p{4.0cm} p{12.0cm}}
\toprule
\textbf{Category} & \textbf{Setting} \\
\midrule
Policy initialization & Qwen3.5-9B policy initialized from the ReasonEdit SFT checkpoint \\
\midrule
Adapters & LoRA rank $32$, alpha $64$, dropout $0.05$ \\
\midrule
Prompt data & Deduplicated source image, edited image, and instruction triplets \\
\midrule
Batching & per-device train batch size $1$; gradient accumulation $4$; per-device eval batch size $1$ \\
\midrule
Rollouts & $G=4$ generations per training prompt; $1$ generation per evaluation prompt \\
\midrule
Sampling & temperature $0.9$; top-$p$ $0.95$ \\
\midrule
Sequence limits & maximum prompt length $3072$; maximum completion length $768$ \\
\midrule
Optimization & learning rate $1\times10^{-5}$; weight decay $0.0$; warmup ratio $0.03$; cosine scheduler \\
\midrule
RL regularization & reference-model KL coefficient $\beta=0.04$ \\
\midrule
Reward & Remote RE-Reward server with $0.3/0.4/0.3$ weights for logicality, accuracy, and usefulness \\
\bottomrule
\end{tabular}}
\end{table*}

\section{Model comparison and experimental details}
\label{sec:supp_comparison}

\subsection{Data splits}
For all experiments, we strictly separated the training/validation sets from the test sets. ReasonEdit-CoT-12K serves as a subset with text evaluations derived from EBench-18K\cite{lmm4edit}; consequently, we strictly aligned our training and test splits with the original sample partitioning of EBench-18K. For ReasonEdit-Reward-113K, we employed hash-based deduplication sampling. This ensures that all edited images and evaluation texts derived from a single source image do not simultaneously appear in both the training and test sets, thereby preventing data leakage. For all zero-shot datasets, we did not perform any fine-tuning on the datasets themselves. Instead, we directly conducted performance testing using the provided benchmarks.

\subsection{Compared reward and evaluation methods}
The main paper compares ReasonEdit and RE-Reward against several categories of methods:
\begin{itemize}[left=12pt, labelsep=0.6em, labelwidth=0pt]
\item \textbf{Traditional full-reference IQA metrics}: SSIM \cite{SSIM}, SC-SSIM \cite{SCSSIM}, GMSD \cite{GMSD}, and related measures compare low-level similarity between reference and distorted images.
\item \textbf{Traditional no-reference IQA metrics}: DIIVINE \cite{DIIVINE}, BRISQUE \cite{BRISQUE}, and NIQE \cite{NIQE} estimate perceptual quality without a reference image.
\item \textbf{Deep IQA metrics}: LPIPS \cite{LPIPS}, CVRKD \cite{CVRKD}, AHIQ \cite{AHIQ}, MANIQA \cite{MANIQA}, TOPIQ \cite{TOPIQ}, and Q-Align \cite{qalign} use learned visual representations for perceptual or quality assessment.
\item \textbf{Vision-language reward metrics}: CLIPScore \cite{clipscore}, ImageReward \cite{imagereward}, PickScore \cite{pickscore}, and VQAScore \cite{vqa} provide text-image or preference-style scoring signals.
\item \textbf{MLLM evaluators}: Qwen-family, InternVL-family, DeepSeek-VL-family, Gemini-family, and GPT-family models are evaluated as zero-shot or prompted multimodal judges where applicable \cite{qwen3,internvl3_5,deepseekvl2,gemini2,gemini3,gpt5}.
\item \textbf{Image-editing-specific evaluators}: LMM4Edit \cite{lmm4edit}, EditReward \cite{editreward}, and EditScore \cite{editscore} are directly relevant because they target image-editing evaluation rather than generic image quality alone.
\end{itemize}

\subsection{Benchmark reporting protocol}
\label{sec:supp_benchmark_protocol}
Table~\ref{tab:supp_benchmark_protocol} summarizes how the main paper reports each benchmark.

\begin{table}[t]
\centering
\caption{Evaluation protocol by benchmark.}
\label{tab:supp_benchmark_protocol}
\small
\setlength{\tabcolsep}{6pt}
\begin{tabular}{p{3.5cm} p{9.5cm}}
\toprule
\textbf{Benchmark} & \textbf{Reported metric in this paper} \\
\midrule
ReasonEdit-Reward-113K & SRCC, KRCC, and PLCC for logicality, accuracy, and usefulness. \\
\midrule
EBench-18K \cite{lmm4edit} & SRCC, KRCC, and PLCC for visual quality, instruction alignment, and content preservation. \\
\midrule
GenAI-Bench \cite{genai} & Pairwise accuracy on image-editing preferences. \\
\midrule
AURORA-Bench \cite{aurora} & Pairwise accuracy on public image-editing preferences. \\
\midrule
EditScore-Bench \cite{editscore} & Pairwise accuracy for reward-style image-editing evaluation. \\
\midrule
ImagenHub \cite{imagenhub} & SRCC after mapping the three predicted score dimensions to an overall score. \\
\midrule
IEQA \cite{xu2026edithf1mmillionscalerichhuman} & Average SRCC over matched scoring dimensions. \\
\bottomrule
\end{tabular}
\end{table}

\subsection{Prompting protocol for zero-shot MLLM baselines}
For MLLM baselines used as zero-shot judges, the prompt should request structured reasoning before final scores. For RE-Reward evaluation, the final response is converted into logicality, accuracy, and usefulness scores. For image-editing quality evaluation, the final response is converted into visual quality, instruction alignment, and content preservation scores. 

Here is the prompt for \textbf{zero-shot reward} tests:

\begin{tcolorbox}[title=Prompt Template, colback=gray!5, colframe=black]
You are an expert in critique text quality assessment for image editing tasks. Output only the required JSON without additional text. \\
Original Image: \verb+\img1+ \\
Edited Image: \verb+\img2+ \\
Edit Instruction: \verb+<instruction_text>+ \\
Evaluation Framework:\\
\textbf{\lbrack Logicality Rubric\rbrack}\\
1 = Confused reasoning, severe contradictions, or failure to understand instruction/edit.\\
3 = Weak reasoning, partial misunderstanding, or mild logical jumps.\\
5 = Extremely clear, well-structured, and logically airtight.\\
\textbf{\lbrack Accuracy Rubric\rbrack}\\
1 = Severe factual errors, visual/compliance hallucinations, or unsupported praise.\\
3 = Mostly correct but misses fine-grained issues.\\
5 = Fully accurate, hallucination-free, covers all relevant facts.\\
\textbf{\lbrack Usefulness Rubric\rbrack}\\
1 = Useless, generic, or unsupported praise.\\
3 = Contains some useful evaluation but misses key information.\\
5 = Highly valuable, evidence-based, explains critical failures/successes.\\
Scoring Rules:\\
- Judge critique text only; never assume edits exist based on instruction.\\
- Penalize hallucinated objects/compliance and unsupported praise.\\
- Assign 2/4 for borderline cases.\\
- Output exactly three JSON scores (1-5) with one-sentence English reasons.\\
\textbf{\lbrack Final Output\rbrack}\\
After [Final Output], provide JSON in this exact format:\\
\{"logicality": \{"score": X, "reason": "Short factual sentence."\}, "accuracy": \{"score": X, "reason": "Short factual sentence."\}, "usefulness": \{"score": X, "reason": "Short factual sentence."\}\}
\end{tcolorbox}

\subsection{Additional results}
\begin{figure}[htbp]
    \centering
    \includegraphics[width=\linewidth]{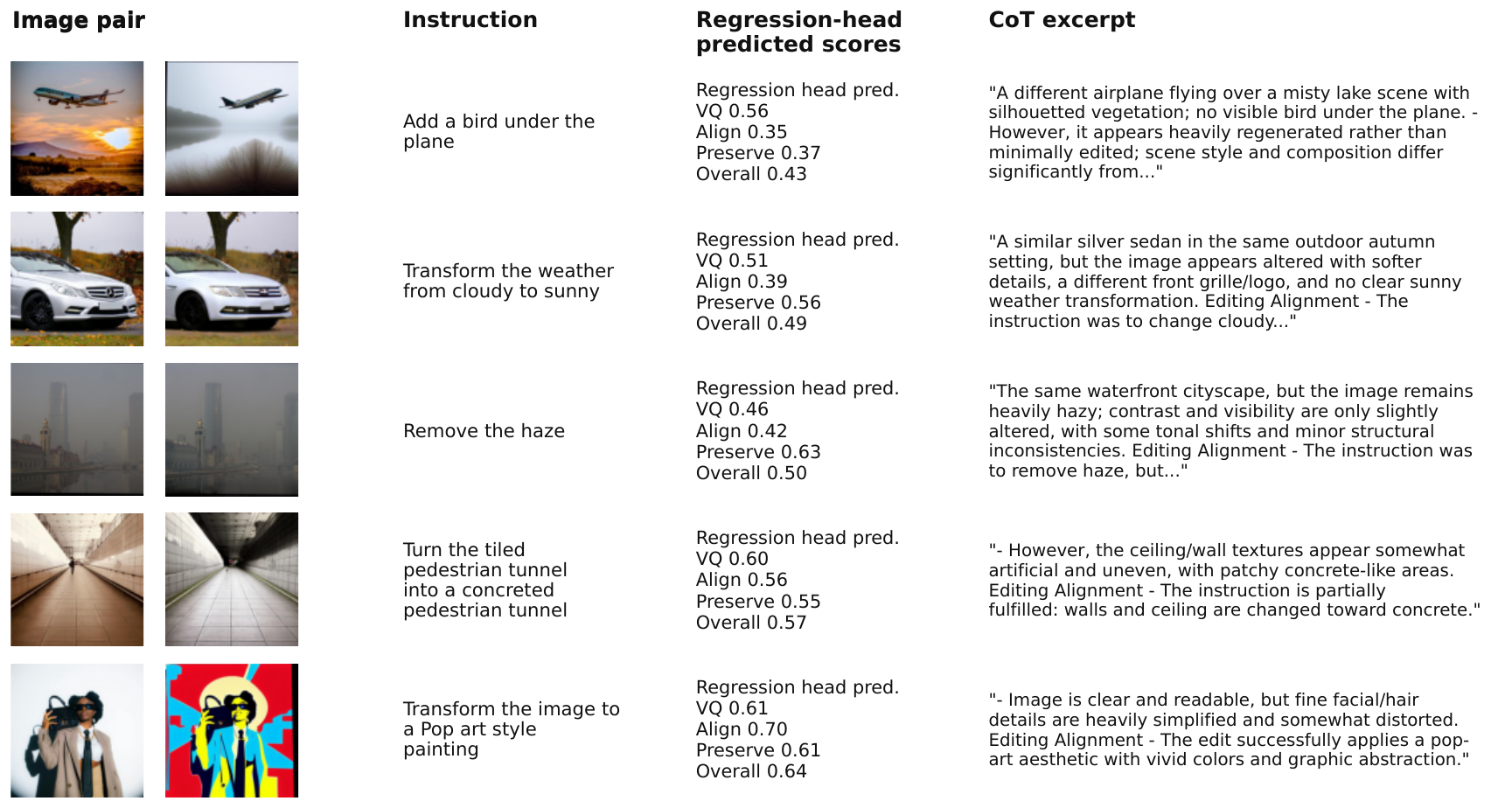}
    \caption{Example where interpretable reasoning reveals a failure not captured by a single scalar score}
    \label{fig:supp_disagreement_case}
\end{figure}

This section include examples illustrating how ReasonEdit grounds a score in visible evidence.
Figure~\ref{fig:supp_disagreement_case} reserves a case where scalar metrics and human reasoning diverge.

\section{More ablation and analysis}
\label{sec:supp_ablation}

\subsection{Ablation of score-anchor pooling}
Experiment indicates that the "prefill output" contains most comprehensive signal for the MLP regression head to extract. Another advantages for this is the optional score-only mode, which saves a lot of time and would be useful as an online reward model, with exactly the same performance as CoT enabled. Table~\ref{tab:supp_anchor_ablation} reserves the final ablation for this design choice.

\begin{table}[htbp]
\centering
\caption{The ablation of regression pooling strategy}
\label{tab:supp_anchor_ablation}
\small
\setlength{\tabcolsep}{6pt}
\begin{tabular}{p{3.5cm} p{2.4cm} p{2.4cm} p{3.5cm}}
\toprule
\textbf{Pooling anchor} & \textbf{Format stability} & \textbf{Score SRCC} & \textbf{Notes} \\
\midrule
<SCORE\_ANCHOR> & 98.07\% & 0.907 & \makecell[l]{Unexpected structural\\failure happens frequently} \\
\midrule
\lbrack Final Assessment\rbrack & 99.89\% & 0.915 & - \\
\midrule
\textbf{Prefill output} & \textbf{100\%} & \textbf{0.932} & \makecell[l]{Absolute stable \\ high performance} \\
\bottomrule
\end{tabular}
\end{table}

\section{Compute cost}
\label{supp_cost}
Table~\ref{tab:supp_compute_cost} shows the detailed compute cost.

\begin{table*}[htbp]
\centering
\caption{Compute resources and runtime}
\label{tab:supp_compute_cost}
\setlength{\tabcolsep}{7pt}
\resizebox{\textwidth}{!}{
\begin{tabular}{p{4.0cm} p{4.0cm} p{4.0cm} p{4.0cm}}
\toprule
\textbf{Stage} & \textbf{Hardware} & \textbf{GPU memory} & \textbf{Runtime} \\
\midrule
RE-Reward training & 4*NVIDIA RTX 4090 & 4*24G & 11.4h\\
\midrule
ReasonEdit SFT & 4*NVIDIA RTX 4090 & 4*24G & 14.6h \\
\midrule
ReasonEdit GRPO policy & 4*NVIDIA A6000 & 4*48G & 110h \\
\midrule
Inference & NVIDIA RTX 4090 & 24G & \makecell[l]{1s/sample (score-only)\\20s/sample (CoT)}\\
\bottomrule
\end{tabular}}
\end{table*}

\section{Reproducibility and release details}
\label{sec:supp_reproducibility}

\subsection{Software environment}
The main experiments use PyTorch, Transformers, PEFT, vLLM, and related multimodal preprocessing libraries. The following list shows the detailed environment information used during SFT and RM training.

\begin{itemize}
      \item Driver / CUDA: 550.54.14 / CUDA 12.4
      \item Python: Python 3.10.20
      \item Same:
      \begin{itemize}
          \item torch 2.6.0+cu124
          \item transformers 5.5.3
          \item accelerate 1.13.0
          \item bitsandbytes 0.49.2
          \item numpy 2.2.6
          \item pandas 2.3.3
          \item scipy 1.15.3
          \item timm 1.0.26
          \item safetensors 0.7.0
      \end{itemize}
      \item SFT environments: 
      \begin{itemize}
          \item peft 0.19.0
          \item datasets 4.8.4
          \item trl 1.1.0
          \item deepspeed 0.18.9
          \item huggingface\_hub 1.10.2
      \end{itemize}
      \item RM environments: 
      \begin{itemize}
          \item peft 0.18.1
          \item huggingface\_hub 1.10.1
      \end{itemize}
  \end{itemize}

\subsection{Open access statement}
We will open-source the LoRA weights and inference scripts for ReasonEdit and the reward model RE-Reward upon paper submission. The final curation of the dataset is still in progress, and we will open-source it once completed.

\section{Limitations}
\label{sec:supp_limitations}

\subsection{Dependence on annotation quality}
ReasonEdit and RE-Reward are trained from human ratings and expert-refined CoT texts. Their behavior is therefore bounded by the consistency, coverage, and perceptual assumptions of the annotation pool. Although the annotation pipeline uses calibration, quality control, and distribution-aware normalization, ambiguous instructions and culturally dependent judgments can still produce legitimate disagreement.

\subsection{Dependence on candidate-generator coverage}
ReasonEdit-CoT and ReasonEdit-Reward inherit linguistic and reasoning patterns from the MLLMs used to generate candidate critiques. Human filtering and expert revision reduce this bias but may not remove it completely. If future editing models produce failure modes not represented in the current candidate pool, ReasonEdit may under-explain those cases.

\subsection{Fine-grained visual failures}
Due to constrained RAM in our GPUs, the pixel budget in our training set-ups are limited to 262,144(512*512). Although the inference supports higher pixel budget, this change would lead to distribution shift and slightly damage the evaluation accuracy. Very small text edits, subtle identity changes, localized artifacts, and fine-grained physical inconsistencies remain challenging. These cases may require higher-resolution visual processing, OCR-aware reasoning, or specialized region-level annotations that are not fully covered by the current scalar-plus-CoT training setup.

\subsection{Overly strict in CoT evaluation texts}
Although we have undergone RLHF with GRPO, the model is sometimes too strict with well-edited images in the output CoT texts.

\section{Potential social impact}
\label{sec:supp_impact}

\subsection{Positive impacts}
ReasonEdit can make image-editing evaluation more transparent by explaining why an edit succeeds or fails. This can support content creation, dataset curation, model debugging, benchmark auditing, and human-in-the-loop quality control. The dataset also provides a structured resource for studying how multimodal evaluators reason about visual evidence, instruction following, and preservation of non-edited content.

\subsection{Negative impacts}
More accurate evaluation and critique generation for image editing can indirectly improve the realism and controllability of manipulated images. This creates dual-use risks, including more convincing deceptive edits, misinformation, impersonation, or privacy-sensitive image manipulation. The model may also inherit biases from source data, annotator judgments, and MLLM-generated critiques, which could affect evaluations involving people, cultural symbols, or subjective aesthetics.

\subsection{Safeguards}
\label{supp_safeguards}
\begin{itemize}
    \item \textbf{Content Safety:} We source resources from open-source datasets and public photography websites. These resources are published with the consent of the photographers and subjects involved, ensuring they do not contain any private information, violence, intentional bias, or other harmful content.
    
    \item \textbf{Usage Safety:} Our model requires users to provide two images and a limited text description. This requirement restricts the model's scope of application to evaluating the interpretability of image editing and providing reward signals.
    
    \item \textbf{Generation Safety:} Although the model possesses text generation capabilities, its overall safety is high. The model is trained on a reliable base model that has undergone debiasing and human preference alignment, capabilities that are inherited by our model. Furthermore, the model is fine-tuned on a large volume of fixed-format data and restricted by system prompts, ensuring it outputs evaluation text almost exclusively in a specific format rather than potentially harmful generic text. The model has also demonstrated high resistance to prompt injection attacks.
\end{itemize}

\subsection{Declaration of MLLM usage}
MLLMs are part of the core methodology rather than being used only for writing assistance. They are used to generate candidate CoT critiques, provide baselines for critique and image-editing evaluation, instantiate RE-Reward and ReasonEdit backbones, and provide reward-guided training signals through the GRPO pipeline.



\end{document}